\documentclass[10pt,twocolumn,letterpaper]{article}

\usepackage[pagenumbers]{cvpr} % To force page numbers, e.g. for an arXiv version
\usepackage[normalem]{ulem}
\usepackage[title]{appendix}

\usepackage{graphicx}
\usepackage{amsmath}
\usepackage{amssymb}
\usepackage{booktabs}

\usepackage{array}  % ADDED
\usepackage{multirow}  % ADDED
\usepackage[warn]{textcomp}  % ADDED
\usepackage{gensymb}  % ADDED
\usepackage{booktabs}  % ADDED
\usepackage{flushend}  % ADDED
\usepackage{float}  % ADDED
\usepackage{xcolor}  % ADDED
\usepackage{siunitx}  % ADDED
\usepackage{microtype}  % ADDED
\usepackage{comment} % ADDED
\usepackage[T1]{fontenc} % ADDED

\usepackage{pifont}% http://ctan.org/pkg/pifont
\usepackage{wrapfig}

\definecolor{citecolor}{HTML}{65C3A6}
\usepackage[pagebackref,breaklinks,colorlinks]{hyperref}
\DeclareMathOperator*{\argmin}{argmin} 

\usepackage[capitalize]{cleveref}
\crefname{section}{Sec.}{Secs.}
\Crefname{section}{Section}{Sections}
\Crefname{table}{Table}{Tables}
\crefname{table}{Tab.}{Tabs.}

\def\cvprPaperID{2100} % *** Enter the CVPR Paper ID here
\def\confName{CVPR}
\def\confYear{2023}

\def\eg{\emph{e.g}\onedot}

\def\etc{\emph{etc}\onedot} 
 
\def\etal{\emph{et al}\onedot}

\newcommand{\cutsectionup}{\vspace*{-6pt}} 
\newcommand{\cutsectiondown}{\vspace*{-2pt}}
\newcommand{\cutsubsectionup}{\vspace*{-4pt}}
\newcommand{\cutsubsectiondown}{\vspace*{-4pt}}

\newcommand{\cutparagraphup}{\vspace*{-13pt}}

\newcommand{\cutcaptionup}{\vspace*{-9pt}}
\newcommand{\cutcaptiondown}{\vspace*{-5pt}}

\newcommand{\cuttablecaptionup}{\vspace*{-8pt}}
\newcommand{\cuttablecaptiondown}{\vspace*{-4pt}}
\newcommand{\cutequationup}{\vspace*{-4pt}}

\newcommand{\cutabstractup}{\vspace*{-10pt}}

\newcommand\blfootnote[1]{%
  \begingroup
  \renewcommand\thefootnote{}\footnote{#1}%
  \addtocounter{footnote}{-1}%
  \endgroup
}
\begin{document}

\title{GINA-3D: Learning to Generate Implicit Neural Assets in the Wild}

\author{
Bokui Shen$^{1*}$ \quad
Xinchen Yan$^2$ \quad
Charles R. Qi$^2$ \quad
Mahyar Najibi$^2$ \quad
%Boyang Deng$^2$ \\
Boyang Deng$^{1,2\dagger}$ \\
Leonidas Guibas$^3$ \quad
Yin Zhou$^2$ \quad
Dragomir Anguelov$^2$ \\
$^1$Stanford University, $^2$Waymo LLC, $^3$Google\\
%{\tt\small willshen@stanford.edu,\{xcyan,rqi,najibi\}@waymo.com,bydeng@stanford.edu,}\\ 
%{\tt\small guibas@google.com,\{yinzhou,dragomir\}@waymo.com}
}

%textwidth in inch: \printinunitsof{in}\prntlen{\textwidth}
% columnwidth in inch: \printinunitsof{in}\prntlen{\columnwidth}
\twocolumn[{%
\maketitle
\cutabstractup
\cutabstractup

\begin{center}
    \centering
    \captionsetup{type=figure}
    \includegraphics[width=\textwidth]{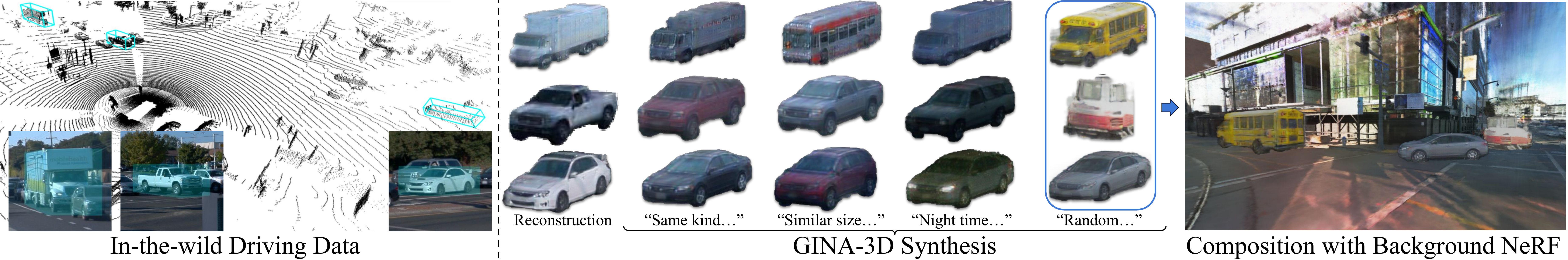}
    \cutcaptionup
    \cutcaptionup
    \captionof{figure}{
    Leveraging in-the-wild data for generative assets modeling embodies a scalable approach for simulation. \textbf{GINA-3D} uses real-world driving data to perform various synthesis tasks for realistic 3D implicit neural assets. 
    Left: Multi-sensor observations in the wild.
    Middle: Asset reconstruction and conditional synthesis.
    Right: Scene composition with background neural fields~\cite{tancik2022block}.
    %Left: conditional synthesis based on an observed scene. Right: unconditional synthesis results, and importing the generated assets to an implicit scene representation of 
    }
    \label{fig:my_label}
\end{center}%
}]

% \twocolumn[{%
% \renewcommand\twocolumn[1][]{#1}%
% \maketitle
% \begin{figure*}
%     \centering
%     \includegraphics[width=\textwidth]{figs/figure_intro.pdf}
%     \cutcaptionup
%     \caption{Caption}
%     \cutcaptiondown
%     \label{fig:my_label}
% \end{figure*}
% }]

%%%%%%%%% ABSTRACT
\begin{abstract}
\cutabstractup
\blfootnote{
$^*$Work done during an internship at Waymo.
$^\dagger$ Work done at Waymo.
}
Modeling the 3D world from sensor data for simulation is a scalable way of developing testing and validation environments for robotic learning problems such as autonomous driving.
However, manually creating or re-creating real-world-like environments is difficult, expensive, and not scalable.
Recent generative model techniques have shown promising progress to address such challenges by learning 3D assets using only plentiful 2D images -- but still suffer limitations as they leverage either human-curated image datasets or renderings from manually-created synthetic 3D environments.
In this paper, we introduce GINA-3D, a generative model that uses real-world driving data from camera and LiDAR sensors to create 
realistic 3D implicit neural assets of diverse vehicles and pedestrians.
Compared to the existing image datasets, the real-world driving setting poses new challenges due to occlusions, lighting-variations and long-tail distributions. 
GINA-3D tackles these challenges by decoupling representation learning and generative modeling into two stages with a learned tri-plane latent structure, inspired by recent advances in generative modeling of images.
To evaluate our approach, we construct a large-scale object-centric dataset containing over 1.2M images of vehicles and pedestrians from the Waymo Open Dataset, and a new set of 80K images of long-tail instances such as construction equipment, garbage trucks, and cable cars. We compare our model with existing approaches and demonstrate that it achieves state-of-the-art performance in quality and diversity for both generated images and geometries.

\end{abstract}
\cutsectionup
\section{Introduction}
\cutsectiondown

Learning to perceive, reason, and interact with the 3D world has been a longstanding challenge in the computer vision and robotics community for decades~\cite{barrow1978recovering,man1982computational,zhang1999shape,tappen2002recovering,hoiem2005automatic,saxena2008make3d,gould2009decomposing,gupta2010blocks}.
Modern robotic systems~\cite{geiger2013vision,maddern20171,chang2019argoverse,caesar2020nuscenes,sun2020scalability,dasari2019robonet,ahn2022can} deployed in the wild are often equipped with multiple sensors (\eg cameras, LiDARs, and Radars) that perceive the 3D environments, followed by an intelligent unit for reasoning and interacting with the complex scene dynamics. 
End-to-end testing and validating these intelligent agents in the real-world environments are difficult and expensive, especially in safety critical and resource constrained domains like autonomous driving.

On the other hand, the use of simulated data has proliferated over the last few years to train and evaluate the intelligent agents under controlled settings~\cite{ros2016synthia,gaidon2016virtual,johnson2017clevr,dosovitskiy2017carla,kolve2017ai2,shah2018airsim,zamir2018taskonomy,abu2018augmented,savva2019habitat,cabon2020virtual,shen2021igibson} in a safe, scalable and verifiable manner.
Such developments were fueled by rapid advances in computer graphics, including rendering frameworks~\cite{blender,juliani2018unity,unrealengine}, physical simulation~\cite{macklin2014unified,coumans2016pybullet} and large-scale open-sourced \textit{asset} repositories~\cite{chang2015shapenet,dai2017scannet,chang2017matterport3d,park2018photoshape,xia2018gibson,mo2019partnet,reizenstein2021common}.
A key concern is to create realistic virtual worlds that align in asset content, composition, and behavior with real distributions, so as to give the practitioner confidence that using such simulations for development and verification can transfer to performance in the real world ~\cite{sadeghi2016cad2rl,muller2018driving,chebotar2019closing,akkaya2019solving,osinski2020simulation,kadian2020sim2real,abeyruwan2022sim2real,manivasagam2020lidarsim,chen2021geosim}. 
However, manual asset creation faces two major obstacles. 
First, manual creation of 3D assets requires dedicated efforts from engineers and artists with 3D domain expertise, which is expensive and difficult to scale~\cite{cabon2020virtual}. Second, real-world distribution
contains diverse examples (including interesting rare cases) and is also constantly evolving~\cite{kar2019meta,varma2019idd}.

% These aforementioned obstacles can potentially be solved by recent development in generative 3D modeling.
Recent developments in the generative 3D modeling offer new perspectives to tackle these aforementioned obstacles, as it allows producing additional realistic but previously unseen examples. 
A sub-class of these approaches, generative 3D-aware image synthesis~\cite{niemeyer2021giraffe,chan2022efficient}, holds significant promise since it enables 3D modeling from partial observations (\eg image projections of the 3D object). 
Moreover, many real-world robotic applications already capture, annotate and update multi-sensor observations at scale.
Such data thus offer an accurate, diverse, task-relevant, and up-to-date representation of the real-world distribution, which the generative model can potentially capture.
However, existing works use either human-curated image datasets with clean observations~\cite{yu2015lsun,yang2015large,karras2017progressive,liu2018large,karras2019style,choi2020stargan} or renderings from synthetic 3D environments~\cite{chang2015shapenet,park2018photoshape}.
Scaling generative 3D-aware image synthesis models to the real world faces several challenges, as many factors are entangled in the partial observations.
First, bridging the in-the-wild images from a simple prior without 3D structures make the learning difficult.
Second, unconstrained occlusions entangle object-of-interest and its surroundings in pixel space, which is hard to disentangle in a purely unsupervised manner.
Lastly, the above challenges are compounded by
a lack of effort in constructing an asset-centric benchmark for sensor data captured in the wild.

In this work, we introduce a 3D-aware generative transformer for implicit neural asset generation, named GINA-3D (\textbf{G}enerative \textbf{I}mplicit \textbf{N}eural \textbf{A}ssets). 
To tackle the real world challenges, we propose a novel 3D-aware Encoder-Decoder framework with a learned structured prior.
Specifically, we embed a tri-plane structure into the latent prior (or \textit{tri-plane latents}) of our generative model, where each entry is parameterized by a discrete representation from a learned codebook~\cite{van2017neural,esser2021taming}. 
The Encoder-Decoder framework is composed of a transformation encoder and a decoder with neural rendering components.
To handle unconstrained occlusions, we explicitly disentangle object pixels from its surrounding with an occlusion-aware composition, using pseudo labels from an off-the-shelf segmenation model~\cite{cheng2020panoptic}.
Finally, the learned prior of tri-plane latents from a discrete codebook can be used to train conditional latents sampling models~\cite{chang2022maskgit}. The same codebook can be readily applied to various conditional synthesis tasks, including object scale, class, semantics, and time-of-day.

To evaluate our model, we construct a large-scale object-centric benchmark from multi-sensor driving data captured in the wild. 
We first extract over 1.2M images of diverse variations for vehicles and pedestrians from Waymo Open Dataset~\cite{sun2020scalability}. 
We then augment the benchmark with long-tail instances from real-world driving scenes, including rare objects like construction equipment, cable cars, school buses and garbage trucks. 
We demonstrate through extensive experiments that GINA-3D outperforms the state-of-the-art 3D-aware generative models, measured by image quality, geometry consistency, and geometry diversity. Moreover, we showcase example applications of various conditional synthesis tasks and shape editing results by leveraging the learned 3D-aware codebook.
The benchmark is publicly available through \href{https://waymo.com/open/data/perception/#object-assets}{waymo.com/open}.

\cutsectionup
\section{Related Work}
\cutsectiondown

We discuss the relevant work on generative 3D-aware image synthesis, 3D shape modeling, and applications in autonomous driving.

\cutparagraphup
\paragraph{Generative 3D-aware image synthesis.}
Learning generative 3D-aware representations from image collections has been increasingly popular for the past decade~\cite{tenenbaum2000separating,reed2014learning,dosovitskiy2015learning,yang2015weakly,kulkarni2015deep,jimenez2016unsupervised,yin2017towards}.
Early work explored image synthesis from disentangled factors such as learned pose embedding~\cite{reed2014learning,yang2015weakly,yin2017towards} or 
compact scene representations
~\cite{dosovitskiy2015learning,kulkarni2015deep}.
Representing the 3D-structure as a compressed embedding, this line of work approached image synthesis by \textit{upsampling} from the embedding space with a stack of 2D deconvolutional layers.
Driven by the progresses in differentiable rendering, there have been efforts~\cite{zhu2018visual,nguyen2019hologan,nguyen2020blockgan,liao2020towards} in baking explicit 3D structures into the generative architectures.
These efforts, however, are often confined to a coarse 3D discretization due to memory consumption.
Moving beyond explicits, more recent work leverages neural radiance fields to learn implicit 3D-aware structures \cite{schwarz2020graf,niemeyer2021giraffe,chan2021pi,hao2021gancraft,zhou2021CIPS3D,gu2022stylenerf,or2022stylesdf,chan2022efficient,poole2022dreamfusion,skorokhodov2022epigraf,deng20233d} for image synthesis.
Schwarz~\etal~\cite{schwarz2020graf} introduced the \textit{Generative Radiance Fields (GRAF)} that disentangles the 3D shape, appearance and camera pose of a single object without occlusions.
Built on top of \textit{GRAF}, Niemeyer~\etal~\cite{niemeyer2021giraffe} proposed the \textit{GIRAFFE} model, which handles scene involving multiple objects by using the compositional 3D scene structure.
Notably, the query operation in the volumetric rendering becomes computationally heavy at higher resolutions.
To tackle this, Chan~\etal~\cite{chan2022efficient} introduced hybrid explicit-implicit 3D representations with \textit{tri-plane} features \textit{(EG3D)}, which showcases image synthesis at higher resolutions. 
Concurrently, \cite{skorokhodov3d} and \cite{sargent2023vq3d} pioneer high-resolution unbounded 3D scene generation on ImageNet using \textit{tri-plane} representations, where ~\cite{sargent2023vq3d} uses a vector-quantized framework and ~\cite{skorokhodov3d} uses a GAN framework. 
Our work is designed for applications in autonomous driving sensor simulation with an emphasis on object-centric modeling.

\cutparagraphup
\paragraph{Generative 3D shape modeling.}
Generative modeling of complete 3D shapes has also been extensively studied, including efforts on synthesizing 3D voxel grids~\cite{wu20153d,wu2016learning,girdhar2016learning,gadelha20173d,smith2017improved,henzler2019escaping,lunz2020inverse,zhou20213d,ibing2021octree}, point clouds~\cite{achlioptas2018learning,yang2019pointflow,mo2020pt2pc}, surface meshes~\cite{sinha2017surfnet,groueix2018papier,pavllo2020convolutional,pavllo2021learning,chen2019learningto,nash2020polygen,gao2022get3d}, shape primitives~\cite{mo2019structurenet,tulsiani2017learning}, and implicit functions or hybrid representations~\cite{liu2019learning,mescheder2019occupancy,luo2021surfgen,chen2019learning,yan2022shapeformer,shen2021deep,mittal2022autosdf,gao2022get3d} using various deep generative models.
Shen~\etal~\cite{shen2021deep} introduced a differentiable explicit surface extraction method called \textit{Deep Marching Tetrahedra (DMTet)} that learns to reconstruct 3D surface meshes with arbitrary topology directly.
Built on top of the EG3D~\cite{chan2022efficient} \textit{tri-plane} features for image synthesis, Gao~\etal~\cite{gao2022get3d} proposed an extension that is capable of generating textured surface meshes using \textit{DMTet} for geometry generation and \textit{tri-plane} features for texture synthesis.
The existing efforts assume access to accurate multi-view silhouettes (often from complete ground-truth 3D shapes) , which does not reflect the real challenges present in data captured in the wild.

\cutparagraphup
\paragraph{Assets modeling in driving simulation.}

Simulated environment modeling has drawn great attention in the autonomous driving domain.
In a nutshell, the problem can be decomposed into asset creation (e.g., dynamic objects and background), scene generation, and rendering.
Early work leverages artist-created objects and background assets to build virtual driving environments~\cite{gaidon2016virtual,dosovitskiy2017carla,richter2016playing} using classic graphics rendering pipelines.
While being able to generate virtual scenes with varying configurations, these methods produce scenes with limited diversity and a significant reality gap.
Many recent works explored different aspects of data-driven simulation, including image synthesis~\cite{hong2018learning,kim2021drivegan,li2019aads,ling2020variational}, assets modeling~\cite{manivasagam2020lidarsim,chen2021geosim,zhang2021ners,zakharov2021single,monnier2022share,muller2022autorf}, scene generation~\cite{kar2019meta,devaranjan2020meta,tan2021scenegen}, and scene rendering~\cite{yang2020surfelgan,tancik2022block,kundu2022panoptic,rematas2022urban}.
In particular, Chen~\etal\cite{chen2021geosim} and Zakharov~\etal\cite{zakharov2021single} performed explicit texture warping or implicit rendering from a single-view observation for each vehicle object. Therefore, their asset reconstruction quality is sensitive to occlusions and bounded by the view angle from a single observation.
Building upon these efforts, more recent work including Muller~\etal\cite{muller2022autorf} and Kundu~\etal\cite{kundu2022panoptic} approached object completion with global or instance-specific latent codes, representing each object asset under the Normalized Object Coordinate Space (NOCS).
In comparison, the latent codes in our proposed model have 3D tri-plane structures which offers several benefits in learning and applications.
More importantly, we can generate previously unseen 3D assets, which is essentially different from object reconstruction.

\cutsectionup
\section{Generative Implicit Neural Assets }
\cutsectiondown

We propose \textbf{GINA-3D}, a scalable framework to acquire 3D assets from in-the-wild data (Sec.~\ref{sec:gina_background}).
Core to our framework is a novel 3D-aware Encoder-Decoder model with a learned structure prior (Sec.~\ref{sec:stage1}).
The learned structure prior can facilitate various downstream applications with an iterative latents sampling model (Sec.~\ref{sec:stage2}) per application.

\subsection{Background.}
\label{sec:gina_background}
\begin{wrapfigure}{r}{0.5\columnwidth}
\centering
    \includegraphics[width=0.46\columnwidth]{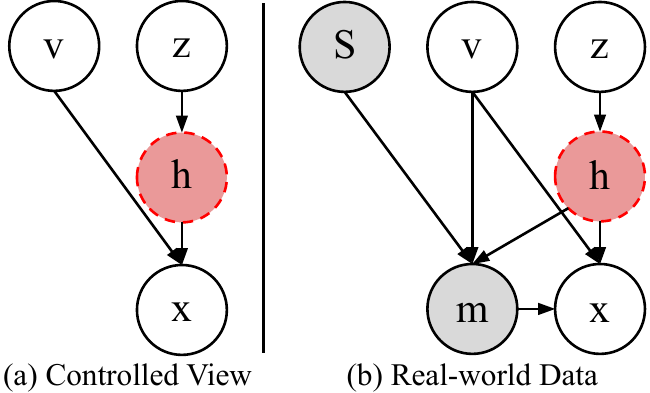}
\cutcaptionup
\caption{Probabilistic Views.}
\cutcaptiondown
\label{fig:prob_graph}
\end{wrapfigure}

\cutsubsectiondown
Given a collection of images containing 3D objects captured in the wild $\mathcal{X} = \{ \mathbf{x}\}$ ($\mathbf{x}$ is an image data sample), 3D-aware image synthesis\cite{tenenbaum2000separating,reed2014learning,dosovitskiy2015learning,yang2015weakly,kulkarni2015deep,jimenez2016unsupervised,yin2017towards,zhu2018visual,nguyen2019hologan,nguyen2020blockgan,liao2020towards,schwarz2020graf,niemeyer2021giraffe,chan2021pi,hao2021gancraft,zhou2021CIPS3D,gu2022stylenerf,or2022stylesdf,chan2022efficient,skorokhodov2022epigraf} aims to learn a distribution of 3D objects.
The core idea is to represent each 3D object as a hidden variable $\mathbf{h}$ within a generative model and further leverage a neural rendering module \texttt{NR} to synthesize a sample image at viewpoint $\mathbf{v}$ through $\mathbf{x} = \texttt{NR}(\mathbf{h}, \mathbf{v})$. 
To model the hidden 3D structure $\mathbf{h}$, the formulation 
introduces a low-dimensional space where latent variables $\mathbf{z}$ (typically a Gaussian) can sample from and connect $\mathbf{h}$ and $\mathbf{z}$ by a generator $\mathbf{h}=f_\theta(\mathbf{z})$, parameterized by $\theta$.
\cutequationup
\begin{align}
\text{Pr}(\mathbf{x}, \mathbf{z} | \mathbf{v}) &= \text{Pr}(\mathbf{x}|\mathbf{z}, \mathbf{v}) \cdot \text{Pr}(\mathbf{z})\label{eqn:joint_prob}
\end{align}
The probabilistic formulation is shown in Fig.~\ref{fig:prob_graph}-a, and Eq.~\ref{eqn:joint_prob}.
Here,
$\text{Pr}(\mathbf{x}|\mathbf{z}, \mathbf{v})$ is the conditional probability of the image given the latent variables and viewpoint, where $\text{Pr}(\mathbf{z})$ and $\text{Pr}(\mathbf{v})$ are the prior distributions.
As the latent variable $\mathbf{z}$ models the 3D objects, one can sample and extract \textit{assets} for downstream applications.
The assets can be either injected into neural representations of scenes~\cite{tancik2022block,kundu2022panoptic}, or transformed into explicit 3D structures such as textured meshes for traditional renders~\cite{dosovitskiy2017carla} or geometry-aware compositing~\cite{chen2021geosim,yang2020surfelgan}.

\setlength{\intextsep}{2pt}%
\setlength{\columnsep}{6pt}%

\cutparagraphup
\paragraph{The challenges in the wild.}
%\noindent \paragraph{The challenges.} 
While human-curated image datasets~\cite{yu2015lsun,yang2015large,karras2017progressive,liu2018large,karras2019style,choi2020stargan} or synthetically generated images with clean background~\cite{chang2015shapenet,park2018photoshape,jimenez2016unsupervised,gao2022get3d} fit into the formulation in Eq.~\ref{eqn:joint_prob}, real-world distributions have unconstrained occlusions due to complex object-scene entanglement.
For example, a moving vehicle can be easily occluded by another object (\eg traffic cones and cars) in an urban driving environment, which further entangle object and scene in the pixel space. 
Moreover, environmental lighting and object diversity lead to a more complex underlying distribution. 

As illustrated Fig.~\ref{fig:prob_graph}-b and Eq.~\ref{eqn:joint_prob_real}, these challenges yield a new probabilistic formulation that the hidden structure $\mathbf{h}$, surrounding scene $\mathbf{S}$ and viewpoint $\mathbf{v}$ jointly contribute to the occlusion ($\mathbf{m}$) and the visible pixels on the object $\mathbf{x}$ through $\mathbf{x} = \texttt{NR}(\mathbf{h}, \mathbf{v}) \odot \mathbf{m}(\mathbf{S},\mathbf{v}, \mathbf{h})$.
\cutequationup
\begin{align}
\text{Pr}(\mathbf{x}, \mathbf{z} | \mathbf{v}, \mathbf{S}) &= \text{Pr}(\mathbf{x}|\mathbf{z}, \mathbf{v}, \mathbf{S}) \cdot \text{Pr}(\mathbf{z}) \label{eqn:joint_prob_real}
\end{align}
Prior art such as GIRAFFE~\cite{niemeyer2021giraffe} tackles the challenges with two assumptions: (1) the scene is composed of a limited number of same-class foreground objects and a background backdrop $\mathbf{S}$; and (2) the real data distribution can be bridged using an one-pass generator $f_\theta (\mathbf{x}; \mathbf{z}, \mathbf{S}, \mathbf{v})$ ($\theta$ is the learned parametrization) conditioned on independently sampled objects $ \mathbf{z}$, scene background $\mathbf{S}$ and the camera viewpoint $\mathbf{v}$ (\eg Multi-variate Gaussian distributions with diagonal variance) through adversarial training.
Unfortunately, the first assumption barely holds for in-the-wild images with unconstrained foreground occlusions.
As shown in Niemeyer~\etal~\cite{niemeyer2021giraffe},
the second assumption can already introduce artifacts due to disentanglement failures.

\cutparagraphup
\paragraph{Our proposal.} 
We focus on interpreting the visible pixels of the object of interest, as synthesizing objects and scene jointly with a generative model is very challenging.
We leverage an auxiliary encoder $E_\phi(\mathbf{x})$ that approximates the posterior $\text{Pr}(\mathbf{z}|\mathbf{x})$ in training the generative model to \textit{reconstruct} the input.
This way, we bypass the need to model complex scene and occlusions explicitly, since paired input and output are now available for supervising the auto-encoding style training.
Specifically, given an image $\mathbf{x}$ and the corresponding occlusion mask $\mathbf{m}$, our objective is to \textit{reconstruct} the visible pixels of the object on the image through $\mathbf{\hat{x}} \odot \mathbf{m}$ where we have the reconstruction $\mathbf{\hat{x}} = \texttt{NR}(G_\theta(\mathbf{z}), \mathbf{v})$ and latent $\mathbf{z} = E_\phi(\mathbf{x})$, respectively.
In practice, we use an off-the-shelf model to obtain the pseudo-labeled object mask as the supervision through $\mathbf{x} \odot \mathbf{m}$.
At the inference time, we can discard the auxiliary encoder $E_\phi$ as our goal is to generate assets from a learned latent distribution (\textit{tri-plane latents} in our case).
To facilitate this, we leverage the vector-quantized formulation~\cite{van2017neural,esser2021taming} to learn a codebook $\mathbb{K} := \{z_n\}_{n=1}^K$ of size $K$ and the mapping from a continuous-valued vector to a discrete codebook entry, where each entry follows a $K$-way categorical distribution.

\begin{figure*}[th]
\centering
\includegraphics[width=\textwidth]{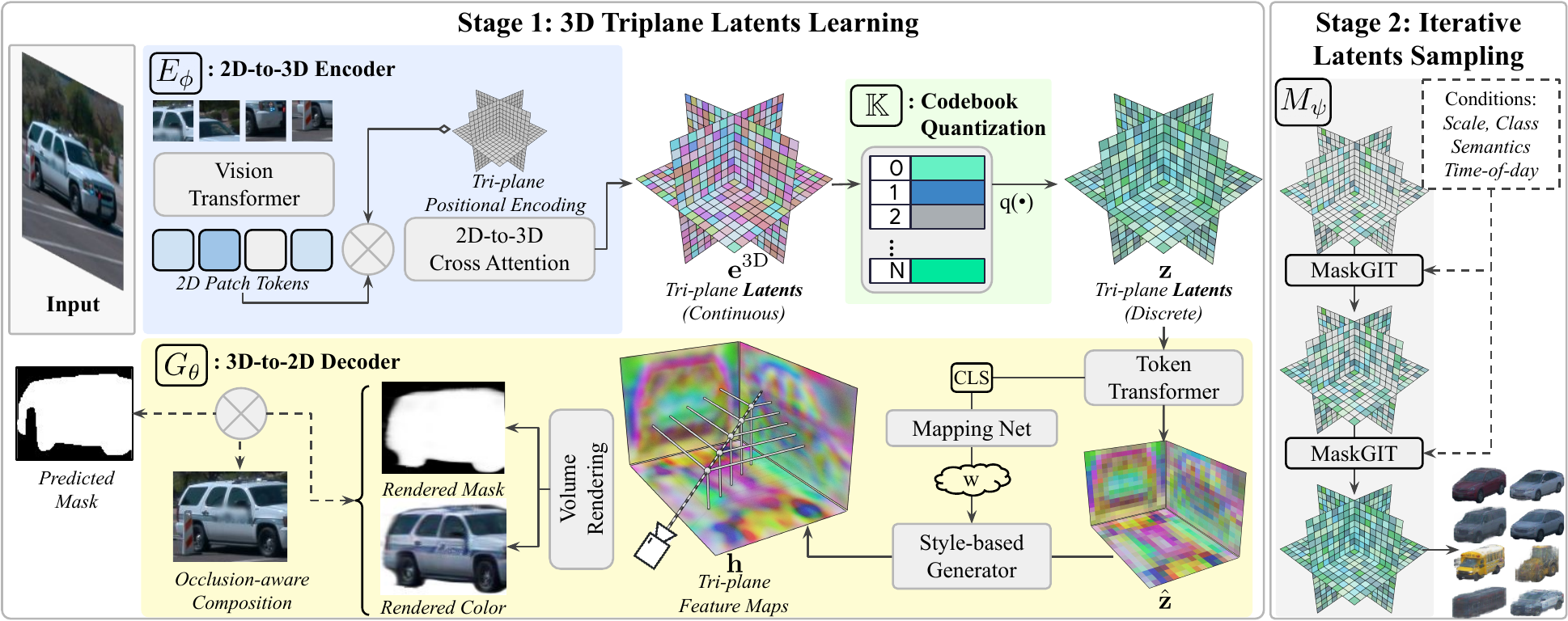}
\cutcaptionup
\cutcaptionup
\caption{We introduce \textbf{GINA-3D}, a 3D-aware generative transformer for implicit neural asset generation. GINA-3D follows a two-stage pipeline, where we learn discrete 3D triplane latents in stage 1 (Sec.~\ref{sec:stage1}) and iterative latents sampling in stage 2 (Sec. ~\ref{sec:stage2}). In stage 1, an input image is first encoded into continuous tri-plane latents $\mathbf{e}^\text{3D}$ using a Transformer-based 2D-to-3D Encoder $E_\phi$. Then, a learnable codebook $\mathbb{K}$ quantize the latents into discrete latents $\mathbf{z}$. Finally, a 3D-to-2D Decoder $G_\theta$ maps $\mathbf{z}$ back to image, using a sequence of Transformer, Style-based Generator and volume rendering. The rendered image is supervised via an occlusion-aware reconstruction loss. In stage 2, we learn iterative latents sampling using MaskGIT~\cite{chang2022maskgit}. Optional conditional information can be used to perform conditional synthesis. The sampled latents can then be decoded into neural assets using the decoder $G_\theta$ learned in stage 1. 
}
\cutcaptiondown
\cutcaptiondown
\label{fig:architecture}
\end{figure*}
% Model architecture. \xcyan{@xcyan: add G, clarify we don't generate background.}
\cutsubsectionup
\subsection{3D Triplane Latents Learning}
\cutsubsectiondown
\label{sec:stage1}
We explain in details the Encoder-Decoder training framework to learn \textit{tri-plane latents} $\mathbf{z}$ (Fig.~\ref{fig:architecture}-left). The framework consists of a 2D-to-3D encoder $E_\phi$, learnable codebook quantization $\mathbb{K}$ and a 3D-to-2D decoder $G_\theta$.

\cutparagraphup
\paragraph{$E_\phi$: 2D-to-3D Encoder.}
We adopt Vision Transformer (ViT)\cite{dosovitskiy2020image} as our image feature extractor that maps $16\times16$ non-overlapping image patches into image tokens of dimension $D_\text{img}$.
%\xcyan{16 $\times$ 16 in the config file.}
%
Since the goal is to infer the latent 3D-structure from a 2D image observation, we associate each image token with tokens in the tri-plane latents using cross-attention modules, which have previously shown strong performance in cross-domain and 2D-to-3D information passing~\cite{hu2021unit,jaegle2021perceiver,sajjadi2022scene,rebain2022attention}.
The cross-attention module uses a learnable tri-plane positional encoding as query, and image patch tokens as key and value. 
The module produces tri-plane embeddings $\mathbf{e}^\text{3D} = E_\phi(\mathbf{x}) \in \mathbb{R}^{N_Z \times N_Z \times 3 \times D_\text{tok}}$, where $D_\text{tok}=32$ and $N_Z = 16$ indicates the dimension of each 3D token and the spatial resolution, respectively.

\cutparagraphup
\paragraph{$\mathbb{K}$: Codebook Quantization for tri-plane latents.}
Given the continuous tri-plane embedding $\mathbf{e}^\text{3D}$, we project it to our $K$-way categorical prior $\mathbb{K}$ through vector quantization. We apply quantization $\mathbf{q}(\cdot)$ of each spatial code $\mathbf{e}^\text{3D}_{ijk} \in \mathbb{R}^{D_\text{tok}}$ on the tri-plane embeddings onto its closest entry $z_n $ in the codebook, which gives tri-plane latents $\mathbf{z}= \mathbf{q}(\mathbf{e}^\text{3D})$. 
\cutequationup
\begin{align}
\mathbf{z}_{ijh} := \Bigl( \argmin_{z_n,n \in \mathbb{K}} \| {\mathbf{e}^\text{3D}_{ijk} - z_n} \|\Bigr) \in \mathbb{R}^{D_\text{tok}}
\label{eqn:vector_quantise}
\end{align}

\cutparagraphup
\paragraph{$G_\theta$: 3D-to-2D Decoder with neural rendering.}
Our decoder takes the tri-plane latents $\mathbf{z}$ as the input and outputs a high-dimensional feature maps $\mathbf{h} \in \mathbb{R}^{N_H \times N_H \times 3 \times D_H}$ used for rendering, where $N_H=256$ and $D_H = 32$ indicates spatial resolution of the tri-plane feature maps and the feature dimension, respectively.
We adopt a token Transformer followed by a Style-based generator~\cite{karras2020analyzing} as our 3D decoder.
The token transformer first produces high-dimensional intermediate features $\hat{\mathbf{z}}\in \mathbb{R}^{N_Z \times N_Z \times 3 \times D_H}$ with an extra \texttt{CLS} token using self-attention modules, which are then feed to the Style-based generator for upsampling.
We use 4 blocks of weight-modulated convolutional layers, each guided by a mapping network conditioned on the \texttt{CLS} token.

Given the feature maps, we use a shallow MLP that takes a 3D point $\mathbf{p}$ and the hidden feature tri-linearly interpolated at the query location $\mathbf{h}(\mathbf{p})$ as input, following~\cite{peng2020convolutional,chan2022efficient,shen2022acid}.
It outputs a density value $\sigma$ and a view-independent color value $\mathbf{c}$.
We perform volume-rendering with the neural radiance field formulation~\cite{mildenhall2021nerf}. 

\cutparagraphup
\paragraph{Training.}
Our framework builds upon the vector-quantized formulations~\cite{van2017neural,razavi2019generating,chen2020generative,esser2021taming,yu2021vector,yu2022scaling,chang2022maskgit,villegas2022phenaki} where we focus on token learning in the first stage.
Specifically, we extend the VQ-GAN training losses, 
where the encoder $E_\phi$, decoder $G_\theta$ and codebook $\mathbb{K}$ are trained jointly with an image discriminator $D$.
As illustrated in Eq.~\ref{eqn:stage1_recon_loss},
we encourage our Encoder-Decoder model to reconstruct the real image $\mathbf{x}$ with $L_2$ reconstruction, LPIPS~\cite{zhang2018unreasonable}, and adversarial loss.
\cutequationup
\begin{align}
&\mathcal{L}_\text{RGB} = \| (\hat{\mathbf{x}} - \mathbf{x}) \odot \mathbf{m}\|^2\nonumber + f^\text{LPIPS}(\hat{\mathbf{x}} \odot \mathbf{m}, \mathbf{x} \odot \mathbf{m}) \\
    %\nonumber \\
    %&\mathcal{L}_\text{GAN} \bigl( \{ G_\theta, \mathcal{Z}\}, D\bigr) = 
    &\mathcal{L}_\text{GAN} = 
    [\log D(\mathbf{x}) + \log (1 - D(\hat{\mathbf{x}})) ]
    \label{eqn:stage1_recon_loss}
\end{align}
To regularize the codebook learning, we apply the latent embedding supervision with a commitment term in Eq.~\ref{eqn:stage1_vq_loss}, where  $\text{sg}[\cdot]$ denotes the stop-gradient operation.
\cutequationup
\begin{align}
    &\mathcal{L}_\text{VQ} = \| \text{sg}[\mathbf{e}^\text{3D}] - \mathbf{z}\|_2^2 + \lambda_\text{commit} \| \text{sg}[\mathbf{z}] - \mathbf{e}^\text{3D} \|_2^2
    \label{eqn:stage1_vq_loss}
\end{align}
We additionally regularize the 3D density field in a weakly supervised manner using the rendered aggregated density (alpha value) $\mathbf{x}_\alpha$, encouraging object pixels to have alpha value 1. % and the non-object region to have density of 0.
To make the loss occlusion aware, we further require  a pixel lies on the non-object region 
to have zero density, inspired by M\"uller~\etal\cite{muller2022autorf}.
This is achieved by restricting the non-object region to cover \texttt{sky} or \texttt{road} class on the pseudo-labeled segmentations (denoted as $\mathbf{m}_{\text{sky,road}}$). 
\cutequationup
\begin{align}
    &\mathcal{L}_\alpha =  \|(\mathbf{x}_\alpha- \mathbf{1}) \odot \mathbf{m} \|^2 + \|\mathbf{x}_\text{alpha} \odot \mathbf{m}_{\text{sky,road}} \|^2 
    \label{eqn:stage1_alpha_loss}
\end{align}
To summarize, we optimize the total objective $\mathcal{L}^*$ in Eq.~\ref{eqn:overall_loss}.
\cutequationup
\begin{align}
    &\mathcal{L}^* = \arg\min_{\phi, \theta, \mathcal{Z}} \max_{D} \mathbb{E}_{\mathbf{x}} \bigl[ \mathcal{L}_\text{VQ} + \mathcal{L}_\text{RGB} + \mathcal{L}_\alpha + \mathcal{L}_\text{GAN}\bigr] 
    \label{eqn:overall_loss}
\end{align}

\cutsubsectionup
\subsection{Iterative Latents Sampling for Neural Assets}
\cutsubsectiondown
\label{sec:stage2}

Once the first stage training is finished, we can now represent neural assets using the learned \textit{tri-plane latents} and \textit{reconstruct} a collection of assets from image inputs.
To generate previously unseen assets with various conditions, we further learn to sample the \textit{tri-plane latents} in the second stage, following the prior works in Generative Transformers~\cite{van2017neural,esser2021taming,yu2021vector,chang2022maskgit}.
More precisely, we transform the quantized embedding $\mathbf{z} \in \mathbb{R}^{N_Z \times N_Z \times 3 \times D_\text{tok}}$ into a discrete sequence $\mathbf{s} \in \{1,...,K\}^{N_Z \times N_Z \times 3}$, where each element corresponds to the index we select from the codebook $\mathbb{K}$ through $\mathbf{s}_{ijk}=n : \mathbf{z}_{ijk}=z_n$.
Following the recent work MaskGIT~\cite{chang2022maskgit}, we use a bidirectional transformer as our latent generator $M_\psi(\mathbf{z})$ that we learn to iteratively sample the latent sequence (Fig.~\ref{fig:architecture}-right).
During training, we learn to predict randomly masked latents $\mathbf{s}_{\bar{M}}$ by minimizing the negative log-likelihood of the masked ones.
\cutequationup
\begin{align}
    \mathcal{L}_{\text{mask}}= - \mathbb{E}_\mathbf{s}[\sum_{\forall ijk : \mathbf{s}_{ijk} = \texttt{[MASK]} } \log \text{Pr}(\mathbf{s}_{ijk} | \mathbf{s}_{\bar{M}})]
\end{align}
At inference time, we iteratively generate and refine latents.
Starting from all latents as $\texttt{[MASK]}$, we iteratively predict all latents simultaneously but only keep the most confident ones in each step. The remaining ones are assigned as $\texttt{[MASK]}$ and the refinement continues.
Finally, the sequence $\mathbf{s}$ can be readily mapped back to neural assets by indexing the codebook $\mathbb{K}$ to generate tri-plane latents $\mathbf{z}$ and decoding using $G_\theta$.
This iterative approach can be applied to asset variations by selectively masking out tokens of a given instance.

\subsection{Expanding Supervision and Conditioning}
\label{sec:augment}
The two-stage training of GINA-3D is flexible in supervision and conditioning.
When we have additional information, we can incorporate it in stage 1 as auxiliary supervision for token learning, or in stage 2 for conditional synthesis.

\begin{comment}
When additional information is present, GINA-3D can efficiently utilize it either in stage 1 as auxiliary supervision, or in stage 2 as conditioned synthesis. We discuss several variants below.
\end{comment}

\cutparagraphup
\paragraph{Unit box vs. Scaled box.}
Object scale information can serve as an additional input to the tri-linear interpolation on the tri-plane feature maps by rescaling the feature maps to span object bounding box (instead of a unit box).

\cutparagraphup
\paragraph{Semantic feature fields.}
Various recent works have demonstrated the effectiveness of learning hybrid representations in the neural rendering~\cite{kobayashi2022decomposing,tschernezki2022neural,song2022nerfplayer} and 2D image synthesis~\cite{zhang2021datasetgan}.
We can naturally incorporate semantic feature fields in our formulation by computing additional channels in our neural rendering MLP. 
We precompute DINO-ViT features~\cite{caron2021emerging} for each image and learn a  semantic feature field to build part correspondence among generated instances.

\cutparagraphup
\paragraph{LiDAR depth supervision.}
When LiDAR point cloud is available in the data, it can be used as the additional supervision through a reconstruction term between the rendered depth and LiDAR depth.

\cutparagraphup
\paragraph{Conditional synthesis.}
Last but not the least, additional information support various applications in conditional synthesis. Denoted as $\mathcal{C}$, it can be fed into our latent prior as $M_\psi(\mathbf{s}_{ijk} | \mathbf{s}_{\bar{M}}, \mathcal{C})$. For example, object scale, object class, time-of-day and object semantic embeddings can also serve as $c$ for control over the generation process. 

\cutsectionup
\section{Experiments}
\cutsectiondown
\begin{table}[b]
    \cutcaptionup
    \cutcaptionup

    \centering
    \begin{tabular}{l|c|c}
        \toprule
         & Images & Unique Instances  \\
        \hline
        WOD-Vehicle & 901K & 23.6K \\
        WOD-Pedestrian & 321K & 8.1K \\
        Longtail-Vehicle & 80K & 3.7K\\
        \bottomrule
    \end{tabular}
    \cutcaptionup
    \caption{Statistics of our object-centric benchmark. Experiments were conducted on a subset with image patches rescaled to $256^2$ resolution.}
    \cutcaptiondown
    \cutcaptiondown
    \label{tab:my_label}
\end{table}

\begin{table*}[th!]
\setlength{\tabcolsep}{5pt}

\centering
\begin{tabular}{ccccccccccccc} 
\toprule
& & & & \multicolumn{4}{c}{Image} 
& \multicolumn{4}{c}{Geometry}   
\\
\cmidrule(lr){5-8} \cmidrule(lr){9-12} 
& & & & \multicolumn{2}{c}{Quality} 
& \multicolumn{2}{c}{Semantic Diversity}
& \multicolumn{2}{c}{Quality}
& \multicolumn{2}{c}{Mesh Diversity}
\\
\cmidrule(lr){5-6} \cmidrule(lr){7-8} \cmidrule(lr){9-10} \cmidrule(lr){11-12} 
\multicolumn{4}{l}{Method}     
% image
& FID\tiny{\textdownarrow}  
%& IS\tiny{\textuparrow}    
& Mask FOU\tiny{\textdownarrow}  
& COV\tiny{\textuparrow}    
& MMD\tiny{\textdownarrow}  
% geometry
%& Consistency\tiny{\textdownarrow} 
& Cons.\tiny{\textdownarrow} 
& Mesh FOU\tiny{\textdownarrow} 
& COV\tiny{\textuparrow} 
& MMD\tiny{\textdownarrow} 
\\ 

\hline

\multicolumn{4}{l}{GIRAFFE~\cite{niemeyer2021giraffe} }
& 105.3
& 43.66 
& 8.24 
& 2.35
& 15.87
& N/A
& N/A 
& N/A   \\
%+mask         &   &    &   &   &    &    &     &     &     &    \\
%\cmidrule(lr){0-0}

%EG3D_WITH_CROP
\multicolumn{4}{l}{EG3D~\cite{chan2022efficient} }
&137.6 
&7.40
&6.26 
&2.37   
&2.38 
&25.7 
&3.12
&4.70 

\\
\hline
%\multicolumn{3}{l}{\textbf{GINA-3D}} \\

& \small{tri-plane $\mathbf{z}$} & \small{scaled box} & \small{LiDAR}

\\ 
%TBD
\cmidrule(lr){2-4}
 \parbox[t]{0mm}{\multirow{4}{*}{\rotatebox[origin=c]{90}{\textbf{GINA-3D}}}}
&$\times$ & $\times$ & $\times$
&147.9
&1.85
&4.78
&2.00
&1.55
& N/A  
&1.95
&2.43
\\

%car_train_dxadv_5_percep0.1.True_TruexTrue.4.f0.10.min0.05_lidar0.0_posenc6_unitboxTrue_11-01_22:10:05
&$\checkmark$ & $\times$ & $\times$
&79.0
&1.82
&19.67
&1.52
&1.27
&11.7
&5.75
&2.21
\\

%GINA-3D Augmented \\
%car_train_dxadv_5_percep0.1.True_TruexTrue.4.f0.10.min0.05_lidar1.0_posenc6_11-01_18:15:24
%+scaled tri-plane $\mathbf{h}$ (\texttt{st}) 
&$\checkmark$ & $\checkmark$ & $\times$
&60.5
&\textbf{1.77}
&20.68
&1.53
&1.06
&\textbf{2.33}   
&8.69
&2.26
\\

%car_train_dxadv_5_percep0.1.True_TruexTrue.4.f0.10.min0.05_lidar1.0_posenc6_unitboxTrue_11-01_22:10:37
&$\checkmark$ & $\checkmark$ & $\checkmark$
&\textbf{59.5}
&1.80
&\textbf{25.00}
&\textbf{1.46}
&\textbf{0.98}
&4.57   
&\textbf{11.42}
&\textbf{2.17} 
\\

% TBD
%+DINO, \texttt{st}
%&63.5
%&1.76
%&20.88
%&1.53
%&1.01
%&3.75   
%&9.18
%&2.24
%\\

\bottomrule
\end{tabular}
\cuttablecaptionup
\caption{Quantitative evaluation on the realism and diversity of generated image and geometry (metrics details in Sec.~\ref{sec:exp_quant}).
%\xcyan{TODO: refine the table.}
}
\cuttablecaptiondown
\cuttablecaptiondown
\cuttablecaptiondown
\cuttablecaptiondown
\label{tab:big-table}
\end{table*}

\subsection{Object-centric Benchmark}
\cutsubsectiondown
We select the Waymo Open Dataset (WOD)~\cite{sun2020scalability} as it is one of the largest and most diverse autonomous driving datasets, containing rich geometric and semantic labels such as object bounding boxes and per-pixel instance masks.
Specifically, the dataset includes 1,150 driving scenes captured mostly in downtown San Francisco and Phoenix, each consisting of 200 frames of multi-sensor observations.
%
%Each data frame includes 3D point clouds from LiDAR sensors and high-resolution images from five cameras (positioned at Front, Front-Left, Front-Right, Side-Left, and Side-Right).
%
%The objects were captured in the wild and their images exhibit large variations due to object interactions (e.g., heavy occlusion and distance to the robotic platform), sensor artifacts (e.g., motion blur and rolling shutter) and environmental factors (e.g., lighting and weather conditions).
%\cutparagraphup
%\paragraph{Object-centric Benchmark.}
To construct an object-centric benchmark, we propose a \textit{coarse-to-fine} procedure to extract collections of single-view 2D photographs by leveraging 3D object boxes, camera-LiDAR synchronization, and fine-grained 2D panoptic labels.
First, we leverage the 3D box annotations to exclude objects beyond certain distances to the surveying vehicle in each data frame (e.g., $40m$ for pedestrians and $60m$ for vehicles, respectively).
At a given frame, we project 3D point clouds within each 3D bounding box to the most visible camera and extract the centering patch to build our single-view 2D image collections.
Furthermore, we train a Panoptic-Deeplab model~\cite{cheng2020panoptic,mei2022waymo} using the 2D panoptic segmentations on the labeled subset and create per-pixel \textit{pseudo-labels} for each camera image on the entire dataset.
This allows us to differentiate pixels belonging to the object of interest, background, and occluder (e.g., standing pole in front of a person).
We further exclude certain patches where objects are heavily occluded using the 2D panoptic predictions.
Even with the filtering criterion applied, we believe that the resulting benchmark is still very challenging due to occlusions, intra-class variations (e.g., truck and sedan), partial observations (e.g., we do not have full 360 degree observations of a single vehicle), and imperfect segmentation.
In particular, we provide accurate registration of camera rays and LiDAR point clouds to the object coordinate frame, taking into account the camera rolling shutter, object motion and ego motion.
We repeat the same process to extract vehicles and pedestrians from WOD, and additional longtail vehicles from our Longtail dataset.
The proposed object-centric benchmark is one of the largest datasets for generative modeling to date, including diverse and longtail examples in the wild.

\begin{figure}
\centering
\includegraphics[width=\columnwidth]{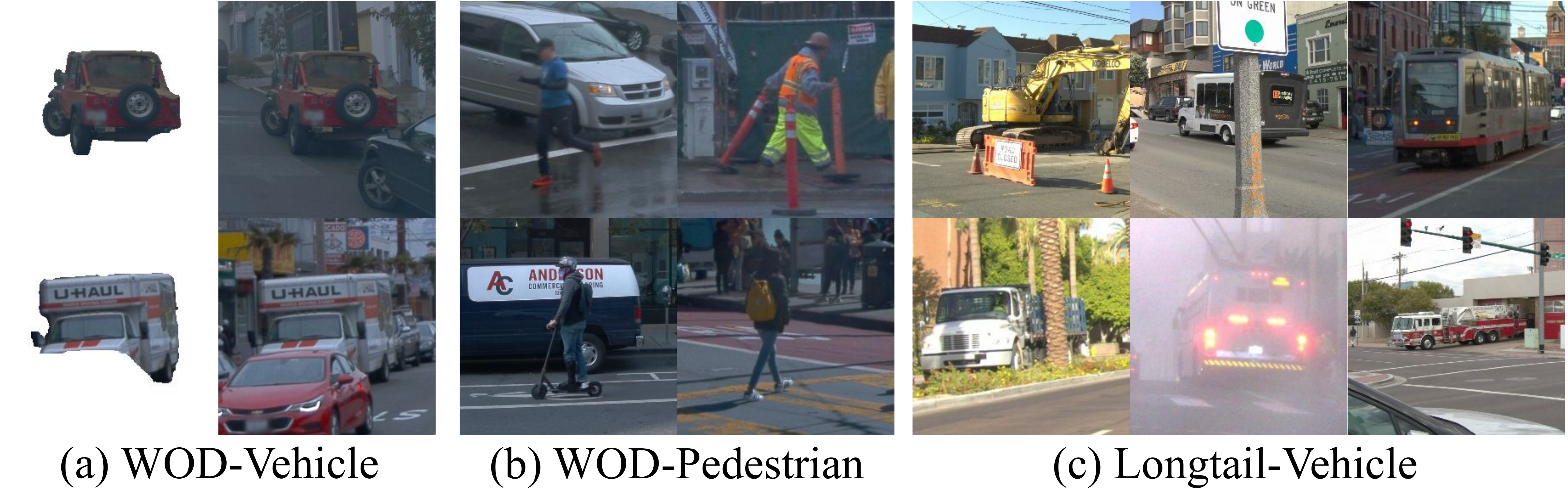}
\cutcaptionup
\cutcaptionup
\caption{Image samples from our object-centric benchmark.}
\cutcaptiondown
\cutcaptiondown
\label{fig:dataset}
\end{figure}

\begin{figure*}
\centering
\includegraphics[width=\textwidth]{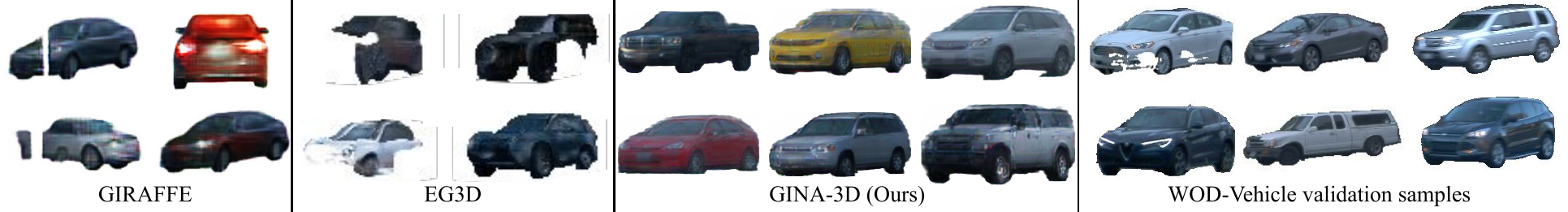}
\cutcaptionup
\cutcaptionup
\caption{Qualitative comparison between GIRAFFE, EG3D and ours with images rendered from a horizontal $30^\circ$ viewpoint. Both baselines fail to disentangle real-world sensor data. GIRAFFE fails to disentangle rotation in object representation, while both baselines fail to disentangle occlusion and produce incomplete shape. We show samples from occlusion-filtered WOD-Vehicle validation set on the right.} 
\cutcaptiondown
\label{fig:baseline_comp}
\end{figure*}

\begin{figure*}
\centering
\includegraphics[width=\textwidth]{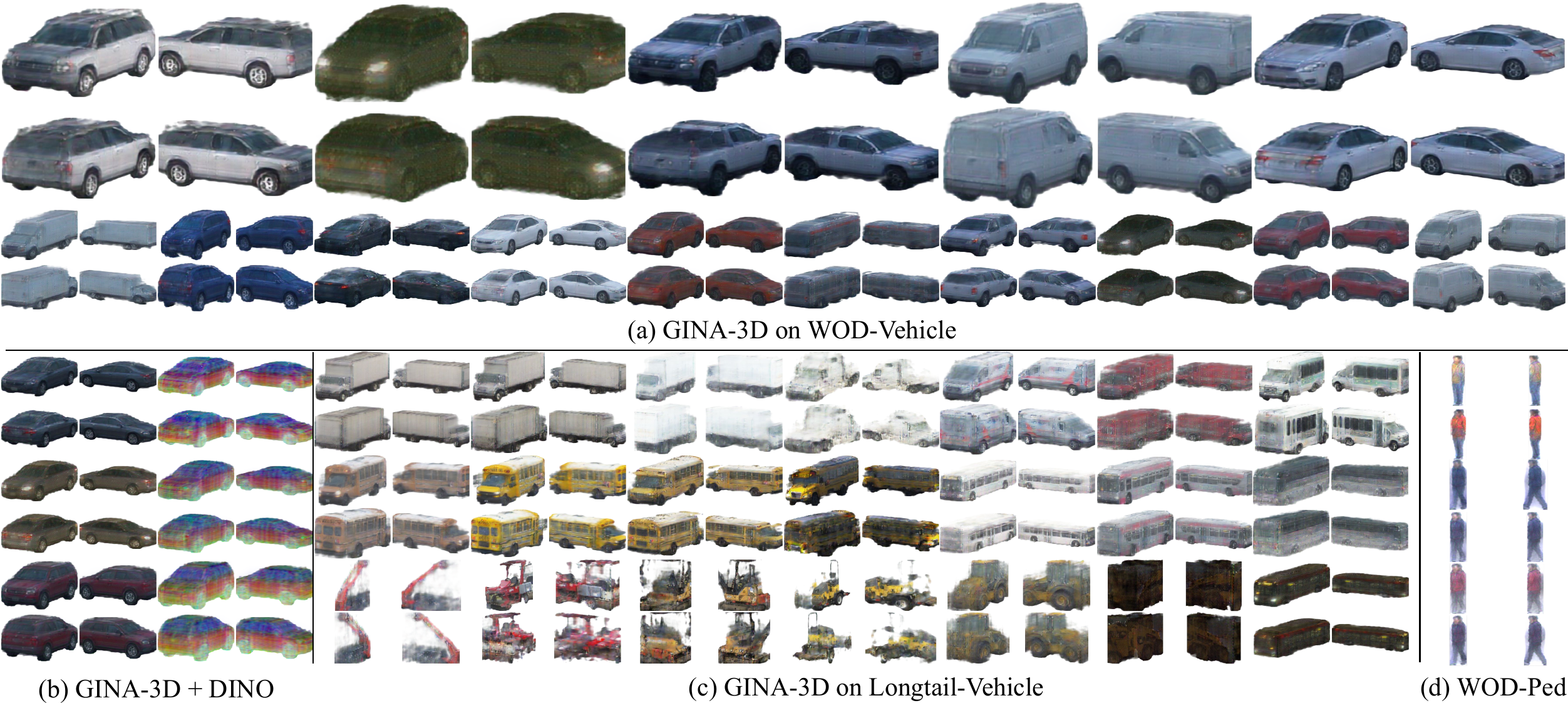}
\cutcaptionup
\cutcaptionup
\caption{Generation from GINA-3D variants. (a) GINA-3D trained on WOD-Vehicle. (b) GINA-3D with additional DINO feature field generation.  (c) GINA-3D trained on Longtail-Vehicle. (d) GINA-3D trained on WOD-Pedestrain. 
}
\cutcaptiondown
\cutcaptiondown
\label{fig:qualitative}
\end{figure*}

\cutsubsectionup
\subsection{Implementation Details}
\cutsubsectiondown
\paragraph{GINA-3D.} 
Our encoder takes in images at resolution of $256^2$ and renders at $128^2$ during training.
Our \textit{tri-plane latents} have a resolution of $16^2$ with a codebook containing 2048 entries and lookup dimension of $32$.
We trained our models on 8 Tesla V100 GPUs using Adam optimizer~\cite{kingma2014adam}, with batch size 32 and 64 in each stage, respectively. We trained stage 1 for 150K steps and stage 2 for 80K steps.  

\cutparagraphup
\paragraph{Baselines.}
We compare against two state-of-the-art methods in the domain, GIRAFFE~\cite{niemeyer2021giraffe} and EG3D~\cite{chan2022efficient}, which we train on our dataset at the resolution of $128^2$.
We noticed that GIRAFFE model trained on full pixels fails to disentangle viewpoints, occlusions and identities.
This makes the extraction of the foreground pixels difficult, as the render mask is only defined at the low dimensional resolution $16^2$.
We instead report the numbers using a model trained by whitening out non-object regions.
For EG3D, we observed that training EG3D with unmasked image leads to training collapse, due to the absence of foreground and background modeling. Thus, we trained EG3D under the same setting.
\cutsubsectionup
\subsection{Evaluations on WOD-Vehicle}
\cutsubsectiondown
\label{sec:exp_quant}
We conduct quantitative evaluations in Table.~\ref{tab:big-table} and visualize qualitative results of different model in Fig.~\ref{fig:baseline_comp}.

\begin{figure*}
    \centering
    \includegraphics[width=\textwidth]{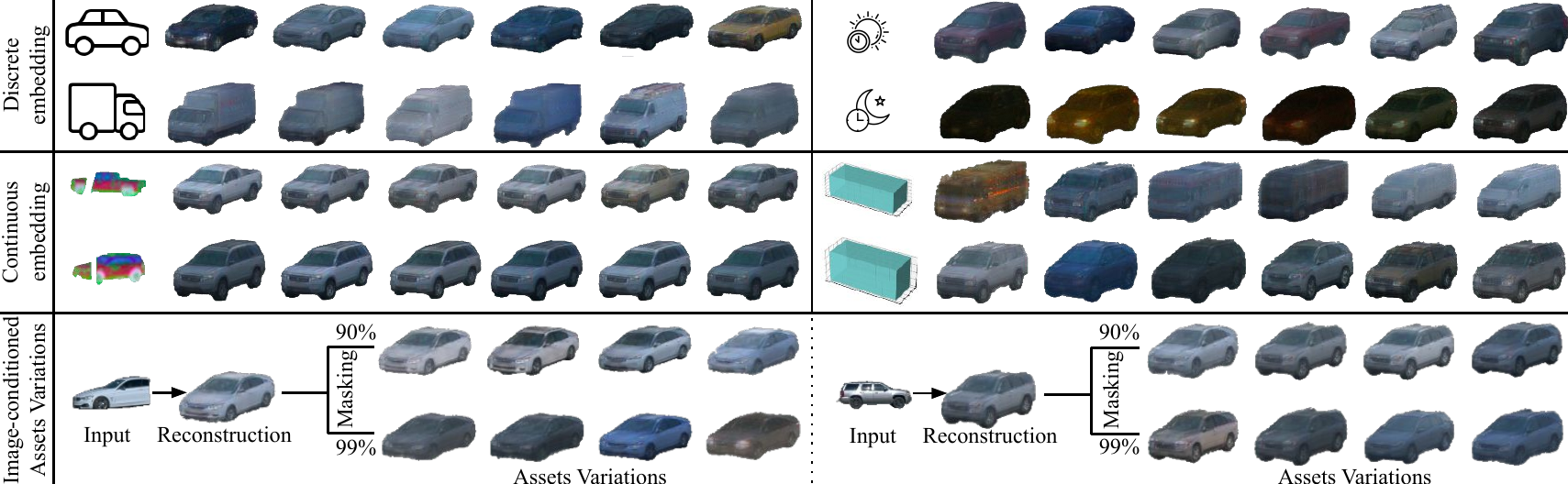}
    \cutcaptionup
    \cutcaptionup
    \caption{GINA-3D unifies a wide range of asset synthesis tasks, all obtained with the same stage 1 decoder and variations of stage 2 training. Top row: Conditional synthesis using discrete conditions (object classes and  time-of-day). 2nd row: Conditional synthesis using continuous conditions (semantic token and object scale). 3rd row: Image-conditioned assets variations by randomizing tri-plane latents. 
    }
    \cutcaptiondown
    \cutcaptiondown
    \cutcaptiondown
    \label{fig:conditional}
\end{figure*}

\cutparagraphup
\paragraph{Image Evaluation.} 
For image quality, we calculate Fr\'echet Inception Distance (FID)~\cite{heusel2017gans} between 50K generated images and all available validation images.
To better reflect the metric on object completeness, we filter images where its object segmentation mask take up at least $50\%$ of the projected 3D bounding box (Fig.\ref{fig:baseline_comp}-right). 
We additionally measure the completeness of the generated images by Mask Floater-Over-Union (Mask FOU), which is defined as the percentage of unconnected pixels over the rendered object region. 
%\xcyan{@will: this is percentage rather than absolute area, right?}
%
To measure the semantic diversity, we compute the Coverage (COV) score and Minimum Matching Distance (MMD)~\cite{achlioptas2018learning} using the CLIP~\cite{radford2021learning} embeddings.
COV measures the fraction of CLIP embeddings in the validation set that has matches in the generated set, and MMD measures the distance between each generated embedding to the closest one in the validation. 
Our model demonstrates significant improvements in FID, image completeness and semantic diversity. 
Without explicit disentanglement, baselines can hardly handle the real distributions, resulting in artifacts of incomplete shapes (Fig.~\ref{fig:baseline_comp}).  

\noindent \textbf{Geometry Evaluation.}
To measure the underlying volume rendering consistency, we follow Or~\etal\cite{or2022stylesdf} and compute the alignment errors between the volume-rendered depth from two viewpoints.
We extract the mesh using marching cubes~\cite{lorensen1987marching} with a density threshold of 10 following EG3D~\cite{chan2022efficient}. 
We measure the completeness by Mesh Floater-Over-Union (Mesh FOU), which is defined as the percentage of the surface area on unconnected mesh pieces over the entire mesh.
Since we do not have ground-truth meshes in the real world data, we approximate mesh diversity by measuring between generated meshes and aggregated LiDAR point clouds within a bounding box from the validation set.
We measure mesh diversity using the aforementioned COV and MMD with a new distance metric.
To account for the incompleteness of LiDAR point clouds, we use a one-way Chamfer distance, which is defined as the mean distance between validation point clouds and their nearest neighbor from a given generated mesh. 
Our model demonstrates significant improvements in volume rendering consistency, shape completeness and shape diversity.

\noindent \textbf{Augmentation and Ablation.}
GINA-3D can naturally incorporate additional supervisions when available.
%
%\xcyan{TODO: Migrate the discussion of DINO to supp.}
We present variations of GINA-3D trained with object scale, LiDAR and DINO~\cite{caron2021emerging} supervision.
With object scale information available, we normalize tri-plane feature maps with the scale on each dimension.
The model trained with rescaled tri-plane resolution yields significant performance boost in both quality and diversity over unit bounding cube, as latents are better utilized.
Moreover, we observe that by adding auxiliary $L_2$ depth supervision from LiDAR, most metrics are improved except Mask and Mesh FOU. 
While LiDAR provides strong signal to underlying geometry, it also introduces inconsistency on transparent surfaces.
We hypothesize that such challenge leads to slightly more floaters, which we leave as future directions to explore.
Alternatively, we can learn additional neural semantic fields through 2D-to-3D feature lifting~\cite{kobayashi2022decomposing}.
By only changing the final layer of the \texttt{NR} MLP, we can learn an additional view-consistent and instance-invariant semantic feature field (Fig.~\ref{fig:qualitative}-b), which can enable future applications of language-conditioned and part-based editing~\cite{gould2009decomposing}
Finally, we perform ablation studies on the key design of tri-plane latents. 
If we remove the tri-plane structure and use a MLP-only \texttt{NR}, the model fails to capture the diversity of real-world data and results in mode collapse, which generates always a mean car shape. 
\cutsubsectionup
\subsection{Applications}
\cutsubsectiondown
\label{sec:exp_app}

\noindent \textbf{Generating long-tail instance.}
Our data-driven framework is scalable to new data.
We provide results on GINA-3D trained on Longtail-Vehicle and WOD-Ped dataset in Fig.~\ref{fig:qualitative}-c,d respectively. 
Without finetuning the architecture on the newly collected data, GINA-3D can readily learn to generate long-tail objects from noisy segmentation masks. 
As shown in Fig.~\ref{fig:qualitative}-c, generation results range from trams, truck to construction equipment of various shapes. 
GINA-3D can also be applied to other categories (e.g. pedestrian, Fig.\ref{fig:qualitative}-d). 
Results show moderate shape and texture diversity.

\noindent \textbf{Conditional synthesis.}
As described in Sec.~\ref{sec:augment}, the flexibility of the two-stage approach makes it a promising candidate for conditional asset synthesis.
Specifically, we freeze the stage 1 model, and train variations of MaskGIT by passing in different conditions.
We provide results for three kinds of conditional synthesis tasks in Fig.~\ref{fig:conditional}, namely discrete embeddings (object class, time-of-day), continuous embeddings, and image-conditioned generation.
For image-conditioned asset reconstruction and variations, we first infer the latents using the encoder model and then sample asset variations by controlling masking ratio of the reconstructed \textit{tri-plane latents}.
The more tokens are masked, the wider the variation range becomes. We provide more details for conditional synthesis in the supplementary material.

\subsection{Limitations}
\cutsubsectiondown
\noindent \textbf{Misaligned 3D bounding boxes.} As in our WOD-Ped results, misaligned boxes lead to mismatch in pixel space, resulting in blurrier results. Latest methods in ray-based~\cite{sajjadi2022scene} or patch-based~\cite{skorokhodov2022epigraf} learning are promising directions.

\noindent \textbf{Few-shot and transfer learning.} Though our data-driven approach achieves reasonable performance by training on Longtail-Vehicle alone, the comparative scarcity of data leads to lower diversity. How to enable few-shot learning or transfer learning remains an open question.

\noindent \textbf{Transcient effects.} Direction-dependent effect can be incorporated in our pipeline. We believe modeling material~\cite{verbin2022ref} together with LiDAR is an interesting direction.

\section{Conclusion}
\cutsectiondown
\cutsectiondown
In this work, we presented GINA-3D, a scalable learning framework to synthesize 3D assets from robotic sensors deployed in the wild.
Core to our framework is a deep encoder-decoder backbone that learns discrete tri-plane latent variables from partially-observed 2D input pixels.
Our backbone is composed of an encoder with cross-attentions, a decoder with tri-plane feature maps, and a neural volumetric rendering module.
We further introduce a latent transformer to generate tri-plane latents with various conditions including bounding box size, time of the day, and semantic features.
To evaluate our framework, we have established a large-scale object-centric benchmark containing diverse vehicles and pedestrians.
Experimental results have demonstrated strong performance on image quality, geometry consistency and geometry diversity over existing methods.
The benchmark is publicly available through \href{https://waymo.com/open/data/perception/#object-assets}{waymo.com/open}.

\cutparagraphup
\paragraph{Acknowledgements:} 
We based our MaskGIT implementation on Chang \etal~\cite{chang2022maskgit}. 
We thank Huiwen Chang for helpful MaskGIT pointers. 
We acknowledge the helpful discussions and support from Qichi Yang and James Guo.
We thank Mathilde Caron for her DINO implementation and helpful pointers. 
We based our GIRAFFE baseline on the reimplementation by Kyle Sargent. 
We thank Golnaz Ghiasi for helpful pointers on segmentation models. 

\newpage
%%%%%%%%% REFERENCES
{\small
\bibliographystyle{unsrt}
\bibliography{egbib}

\begin{thebibliography}{100}

\bibitem{tancik2022block}
Matthew Tancik, Vincent Casser, Xinchen Yan, Sabeek Pradhan, Ben Mildenhall,
  Pratul~P Srinivasan, Jonathan~T Barron, and Henrik Kretzschmar.
\newblock Block-nerf: Scalable large scene neural view synthesis.
\newblock In {\em CVPR}, 2022.

\bibitem{barrow1978recovering}
Harry Barrow, J~Tenenbaum, A~Hanson, and E~Riseman.
\newblock Recovering intrinsic scene characteristics.
\newblock {\em Comput. vis. syst}, 2(3-26):2, 1978.

\bibitem{man1982computational}
D~Man and A~Vision.
\newblock A computational investigation into the human representation and
  processing of visual information.
\newblock {\em WH San Francisco: Freeman and Company, San Francisco}, 1982.

\bibitem{zhang1999shape}
Ruo Zhang, Ping-Sing Tsai, James~Edwin Cryer, and Mubarak Shah.
\newblock Shape-from-shading: a survey.
\newblock {\em IEEE transactions on pattern analysis and machine intelligence},
  21(8):690--706, 1999.

\bibitem{tappen2002recovering}
Marshall Tappen, William Freeman, and Edward Adelson.
\newblock Recovering intrinsic images from a single image.
\newblock {\em NIPS}, 15, 2002.

\bibitem{hoiem2005automatic}
Derek Hoiem, Alexei~A Efros, and Martial Hebert.
\newblock Automatic photo pop-up.
\newblock In {\em ACM SIGGRAPH 2005 Papers}, pages 577--584. 2005.

\bibitem{saxena2008make3d}
Ashutosh Saxena, Min Sun, and Andrew~Y Ng.
\newblock Make3d: Learning 3d scene structure from a single still image.
\newblock {\em IEEE transactions on pattern analysis and machine intelligence},
  31(5):824--840, 2008.

\bibitem{gould2009decomposing}
Stephen Gould, Richard Fulton, and Daphne Koller.
\newblock Decomposing a scene into geometric and semantically consistent
  regions.
\newblock In {\em ICCV}. IEEE, 2009.

\bibitem{gupta2010blocks}
Abhinav Gupta, Alexei~A Efros, and Martial Hebert.
\newblock Blocks world revisited: Image understanding using qualitative
  geometry and mechanics.
\newblock In {\em ECCV}. Springer, 2010.

\bibitem{geiger2013vision}
Andreas Geiger, Philip Lenz, Christoph Stiller, and Raquel Urtasun.
\newblock Vision meets robotics: The kitti dataset.
\newblock {\em The International Journal of Robotics Research},
  32(11):1231--1237, 2013.

\bibitem{maddern20171}
Will Maddern, Geoffrey Pascoe, Chris Linegar, and Paul Newman.
\newblock 1 year, 1000 km: The oxford robotcar dataset.
\newblock {\em The International Journal of Robotics Research}, 36(1):3--15,
  2017.

\bibitem{chang2019argoverse}
Ming-Fang Chang, John Lambert, Patsorn Sangkloy, Jagjeet Singh, Slawomir Bak,
  Andrew Hartnett, De~Wang, Peter Carr, Simon Lucey, Deva Ramanan, et~al.
\newblock Argoverse: 3d tracking and forecasting with rich maps.
\newblock In {\em CVPR}, 2019.

\bibitem{caesar2020nuscenes}
Holger Caesar, Varun Bankiti, Alex~H Lang, Sourabh Vora, Venice~Erin Liong,
  Qiang Xu, Anush Krishnan, Yu~Pan, Giancarlo Baldan, and Oscar Beijbom.
\newblock nuscenes: A multimodal dataset for autonomous driving.
\newblock In {\em CVPR}, 2020.

\bibitem{sun2020scalability}
Pei Sun, Henrik Kretzschmar, Xerxes Dotiwalla, Aurelien Chouard, Vijaysai
  Patnaik, Paul Tsui, James Guo, Yin Zhou, Yuning Chai, Benjamin Caine, et~al.
\newblock Scalability in perception for autonomous driving: Waymo open dataset.
\newblock In {\em CVPR}, 2020.

\bibitem{dasari2019robonet}
Sudeep Dasari, Frederik Ebert, Stephen Tian, Suraj Nair, Bernadette Bucher,
  Karl Schmeckpeper, Siddharth Singh, Sergey Levine, and Chelsea Finn.
\newblock Robonet: Large-scale multi-robot learning.
\newblock {\em arXiv preprint arXiv:1910.11215}, 2019.

\bibitem{ahn2022can}
Michael Ahn, Anthony Brohan, Noah Brown, Yevgen Chebotar, Omar Cortes, Byron
  David, Chelsea Finn, Keerthana Gopalakrishnan, Karol Hausman, Alex Herzog,
  et~al.
\newblock Do as i can, not as i say: Grounding language in robotic affordances.
\newblock {\em arXiv preprint arXiv:2204.01691}, 2022.

\bibitem{ros2016synthia}
German Ros, Laura Sellart, Joanna Materzynska, David Vazquez, and Antonio~M
  Lopez.
\newblock The synthia dataset: A large collection of synthetic images for
  semantic segmentation of urban scenes.
\newblock In {\em CVPR}, 2016.

\bibitem{gaidon2016virtual}
Adrien Gaidon, Qiao Wang, Yohann Cabon, and Eleonora Vig.
\newblock Virtual worlds as proxy for multi-object tracking analysis.
\newblock In {\em CVPR}, 2016.

\bibitem{johnson2017clevr}
Justin Johnson, Bharath Hariharan, Laurens Van Der~Maaten, Li~Fei-Fei,
  C~Lawrence~Zitnick, and Ross Girshick.
\newblock Clevr: A diagnostic dataset for compositional language and elementary
  visual reasoning.
\newblock In {\em CVPR}, 2017.

\bibitem{dosovitskiy2017carla}
Alexey Dosovitskiy, German Ros, Felipe Codevilla, Antonio Lopez, and Vladlen
  Koltun.
\newblock Carla: An open urban driving simulator.
\newblock In {\em Conference on robot learning}, pages 1--16. PMLR, 2017.

\bibitem{kolve2017ai2}
Eric Kolve, Roozbeh Mottaghi, Winson Han, Eli VanderBilt, Luca Weihs, Alvaro
  Herrasti, Daniel Gordon, Yuke Zhu, Abhinav Gupta, and Ali Farhadi.
\newblock Ai2-thor: An interactive 3d environment for visual ai.
\newblock {\em arXiv preprint arXiv:1712.05474}, 2017.

\bibitem{shah2018airsim}
Shital Shah, Debadeepta Dey, Chris Lovett, and Ashish Kapoor.
\newblock Airsim: High-fidelity visual and physical simulation for autonomous
  vehicles.
\newblock In {\em Field and service robotics}, pages 621--635. Springer, 2018.

\bibitem{zamir2018taskonomy}
Amir~R Zamir, Alexander Sax, William Shen, Leonidas~J Guibas, Jitendra Malik,
  and Silvio Savarese.
\newblock Taskonomy: Disentangling task transfer learning.
\newblock In {\em CVPR}, 2018.

\bibitem{abu2018augmented}
Hassan Abu~Alhaija, Siva~Karthik Mustikovela, Lars Mescheder, Andreas Geiger,
  and Carsten Rother.
\newblock Augmented reality meets computer vision: Efficient data generation
  for urban driving scenes.
\newblock {\em International Journal of Computer Vision}, 126(9):961--972,
  2018.

\bibitem{savva2019habitat}
Manolis Savva, Abhishek Kadian, Oleksandr Maksymets, Yili Zhao, Erik Wijmans,
  Bhavana Jain, Julian Straub, Jia Liu, Vladlen Koltun, Jitendra Malik, et~al.
\newblock Habitat: A platform for embodied ai research.
\newblock In {\em ICCV}, 2019.

\bibitem{cabon2020virtual}
Yohann Cabon, Naila Murray, and Martin Humenberger.
\newblock Virtual kitti 2.
\newblock {\em arXiv preprint arXiv:2001.10773}, 2020.

\bibitem{shen2021igibson}
Bokui Shen, Fei Xia, Chengshu Li, Roberto Mart{\'\i}n-Mart{\'\i}n, Linxi Fan,
  Guanzhi Wang, Claudia P{\'e}rez-D’Arpino, Shyamal Buch, Sanjana Srivastava,
  Lyne Tchapmi, et~al.
\newblock igibson 1.0: a simulation environment for interactive tasks in large
  realistic scenes.
\newblock In {\em IROS}, 2021.

\bibitem{blender}
Blender~Online Community.
\newblock Blender - a 3d modelling and rendering package, 2018.

\bibitem{juliani2018unity}
Arthur Juliani, Vincent-Pierre Berges, Ervin Teng, Andrew Cohen, Jonathan
  Harper, Chris Elion, Chris Goy, Yuan Gao, Hunter Henry, Marwan Mattar, et~al.
\newblock Unity: A general platform for intelligent agents.
\newblock {\em arXiv preprint arXiv:1809.02627}, 2018.

\bibitem{unrealengine}
{Epic Games}.
\newblock Unreal engine.

\bibitem{macklin2014unified}
Miles Macklin, Matthias M{\"u}ller, Nuttapong Chentanez, and Tae-Yong Kim.
\newblock Unified particle physics for real-time applications.
\newblock {\em ACM Transactions on Graphics (TOG)}, 33(4):1--12, 2014.

\bibitem{coumans2016pybullet}
Erwin Coumans and Yunfei Bai.
\newblock Pybullet, a python module for physics simulation for games, robotics
  and machine learning.
\newblock 2016.

\bibitem{chang2015shapenet}
Angel~X Chang, Thomas Funkhouser, Leonidas Guibas, Pat Hanrahan, Qixing Huang,
  Zimo Li, Silvio Savarese, Manolis Savva, Shuran Song, Hao Su, et~al.
\newblock Shapenet: An information-rich 3d model repository.
\newblock {\em arXiv preprint arXiv:1512.03012}, 2015.

\bibitem{dai2017scannet}
Angela Dai, Angel~X Chang, Manolis Savva, Maciej Halber, Thomas Funkhouser, and
  Matthias Nie{\ss}ner.
\newblock Scannet: Richly-annotated 3d reconstructions of indoor scenes.
\newblock In {\em CVPR}, 2017.

\bibitem{chang2017matterport3d}
Angel Chang, Angela Dai, Thomas Funkhouser, Maciej Halber, Matthias Niessner,
  Manolis Savva, Shuran Song, Andy Zeng, and Yinda Zhang.
\newblock Matterport3d: Learning from rgb-d data in indoor environments.
\newblock {\em arXiv preprint arXiv:1709.06158}, 2017.

\bibitem{park2018photoshape}
Keunhong Park, Konstantinos Rematas, Ali Farhadi, and Steven~M Seitz.
\newblock Photoshape: Photorealistic materials for large-scale shape
  collections.
\newblock {\em arXiv preprint arXiv:1809.09761}, 2018.

\bibitem{xia2018gibson}
Fei Xia, Amir~R Zamir, Zhiyang He, Alexander Sax, Jitendra Malik, and Silvio
  Savarese.
\newblock Gibson env: Real-world perception for embodied agents.
\newblock In {\em CVPR}, 2018.

\bibitem{mo2019partnet}
Kaichun Mo, Shilin Zhu, Angel~X Chang, Li~Yi, Subarna Tripathi, Leonidas~J
  Guibas, and Hao Su.
\newblock Partnet: A large-scale benchmark for fine-grained and hierarchical
  part-level 3d object understanding.
\newblock In {\em CVPR}, 2019.

\bibitem{reizenstein2021common}
Jeremy Reizenstein, Roman Shapovalov, Philipp Henzler, Luca Sbordone, Patrick
  Labatut, and David Novotny.
\newblock Common objects in 3d: Large-scale learning and evaluation of
  real-life 3d category reconstruction.
\newblock In {\em ICCV}, 2021.

\bibitem{sadeghi2016cad2rl}
Fereshteh Sadeghi and Sergey Levine.
\newblock Cad2rl: Real single-image flight without a single real image.
\newblock {\em arXiv preprint arXiv:1611.04201}, 2016.

\bibitem{muller2018driving}
Matthias M{\"u}ller, Alexey Dosovitskiy, Bernard Ghanem, and Vladlen Koltun.
\newblock Driving policy transfer via modularity and abstraction.
\newblock {\em arXiv preprint arXiv:1804.09364}, 2018.

\bibitem{chebotar2019closing}
Yevgen Chebotar, Ankur Handa, Viktor Makoviychuk, Miles Macklin, Jan Issac,
  Nathan Ratliff, and Dieter Fox.
\newblock Closing the sim-to-real loop: Adapting simulation randomization with
  real world experience.
\newblock In {\em ICRA}, 2019.

\bibitem{akkaya2019solving}
Ilge Akkaya, Marcin Andrychowicz, Maciek Chociej, Mateusz Litwin, Bob McGrew,
  Arthur Petron, Alex Paino, Matthias Plappert, Glenn Powell, Raphael Ribas,
  et~al.
\newblock Solving rubik's cube with a robot hand.
\newblock {\em arXiv preprint arXiv:1910.07113}, 2019.

\bibitem{osinski2020simulation}
B{\l}a{\.z}ej Osi{\'n}ski, Adam Jakubowski, Pawe{\l} Zi{\k{e}}cina, Piotr
  Mi{\l}o{\'s}, Christopher Galias, Silviu Homoceanu, and Henryk Michalewski.
\newblock Simulation-based reinforcement learning for real-world autonomous
  driving.
\newblock In {\em ICRA}, 2020.

\bibitem{kadian2020sim2real}
Abhishek Kadian, Joanne Truong, Aaron Gokaslan, Alexander Clegg, Erik Wijmans,
  Stefan Lee, Manolis Savva, Sonia Chernova, and Dhruv Batra.
\newblock Sim2real predictivity: Does evaluation in simulation predict
  real-world performance?
\newblock {\em IEEE Robotics and Automation Letters}, 5(4):6670--6677, 2020.

\bibitem{abeyruwan2022sim2real}
Saminda Abeyruwan, Laura Graesser, David~B D'Ambrosio, Avi Singh, Anish
  Shankar, Alex Bewley, and Pannag~R Sanketi.
\newblock i-sim2real: Reinforcement learning of robotic policies in tight
  human-robot interaction loops.
\newblock {\em arXiv preprint arXiv:2207.06572}, 2022.

\bibitem{manivasagam2020lidarsim}
Sivabalan Manivasagam, Shenlong Wang, Kelvin Wong, Wenyuan Zeng, Mikita
  Sazanovich, Shuhan Tan, Bin Yang, Wei-Chiu Ma, and Raquel Urtasun.
\newblock Lidarsim: Realistic lidar simulation by leveraging the real world.
\newblock In {\em CVPR}, 2020.

\bibitem{chen2021geosim}
Yun Chen, Frieda Rong, Shivam Duggal, Shenlong Wang, Xinchen Yan, Sivabalan
  Manivasagam, Shangjie Xue, Ersin Yumer, and Raquel Urtasun.
\newblock Geosim: Realistic video simulation via geometry-aware composition for
  self-driving.
\newblock In {\em CVPR}, 2021.

\bibitem{kar2019meta}
Amlan Kar, Aayush Prakash, Ming-Yu Liu, Eric Cameracci, Justin Yuan, Matt
  Rusiniak, David Acuna, Antonio Torralba, and Sanja Fidler.
\newblock Meta-sim: Learning to generate synthetic datasets.
\newblock In {\em ICCV}, 2019.

\bibitem{varma2019idd}
Girish Varma, Anbumani Subramanian, Anoop Namboodiri, Manmohan Chandraker, and
  CV~Jawahar.
\newblock Idd: A dataset for exploring problems of autonomous navigation in
  unconstrained environments.
\newblock In {\em WACV}. IEEE, 2019.

\bibitem{niemeyer2021giraffe}
Michael Niemeyer and Andreas Geiger.
\newblock Giraffe: Representing scenes as compositional generative neural
  feature fields.
\newblock In {\em CVPR}, 2021.

\bibitem{chan2022efficient}
Eric~R Chan, Connor~Z Lin, Matthew~A Chan, Koki Nagano, Boxiao Pan, Shalini
  De~Mello, Orazio Gallo, Leonidas~J Guibas, Jonathan Tremblay, Sameh Khamis,
  et~al.
\newblock Efficient geometry-aware 3d generative adversarial networks.
\newblock In {\em CVPR}, 2022.

\bibitem{yu2015lsun}
Fisher Yu, Ari Seff, Yinda Zhang, Shuran Song, Thomas Funkhouser, and Jianxiong
  Xiao.
\newblock Lsun: Construction of a large-scale image dataset using deep learning
  with humans in the loop.
\newblock {\em arXiv preprint arXiv:1506.03365}, 2015.

\bibitem{yang2015large}
Linjie Yang, Ping Luo, Chen Change~Loy, and Xiaoou Tang.
\newblock A large-scale car dataset for fine-grained categorization and
  verification.
\newblock In {\em CVPR}, 2015.

\bibitem{karras2017progressive}
Tero Karras, Timo Aila, Samuli Laine, and Jaakko Lehtinen.
\newblock Progressive growing of gans for improved quality, stability, and
  variation.
\newblock {\em arXiv preprint arXiv:1710.10196}, 2017.

\bibitem{liu2018large}
Ziwei Liu, Ping Luo, Xiaogang Wang, and Xiaoou Tang.
\newblock Large-scale celebfaces attributes (celeba) dataset.
\newblock {\em Retrieved August}, 15(2018):11, 2018.

\bibitem{karras2019style}
Tero Karras, Samuli Laine, and Timo Aila.
\newblock A style-based generator architecture for generative adversarial
  networks.
\newblock In {\em CVPR}, 2019.

\bibitem{choi2020stargan}
Yunjey Choi, Youngjung Uh, Jaejun Yoo, and Jung-Woo Ha.
\newblock Stargan v2: Diverse image synthesis for multiple domains.
\newblock In {\em CVPR}, 2020.

\bibitem{van2017neural}
Aaron Van Den~Oord, Oriol Vinyals, et~al.
\newblock Neural discrete representation learning.
\newblock In {\em NeurIPS}, 2017.

\bibitem{esser2021taming}
Patrick Esser, Robin Rombach, and Bjorn Ommer.
\newblock Taming transformers for high-resolution image synthesis.
\newblock In {\em CVPR}, 2021.

\bibitem{cheng2020panoptic}
Bowen Cheng, Maxwell~D Collins, Yukun Zhu, Ting Liu, Thomas~S Huang, Hartwig
  Adam, and Liang-Chieh Chen.
\newblock Panoptic-deeplab: A simple, strong, and fast baseline for bottom-up
  panoptic segmentation.
\newblock In {\em Proceedings of the IEEE/CVF conference on computer vision and
  pattern recognition}, pages 12475--12485, 2020.

\bibitem{chang2022maskgit}
Huiwen Chang, Han Zhang, Lu~Jiang, Ce~Liu, and William~T Freeman.
\newblock Maskgit: Masked generative image transformer.
\newblock In {\em CVPR}, 2022.

\bibitem{tenenbaum2000separating}
Joshua~B Tenenbaum and William~T Freeman.
\newblock Separating style and content with bilinear models.
\newblock {\em Neural computation}, 12(6):1247--1283, 2000.

\bibitem{reed2014learning}
Scott Reed, Kihyuk Sohn, Yuting Zhang, and Honglak Lee.
\newblock Learning to disentangle factors of variation with manifold
  interaction.
\newblock In {\em ICML}. PMLR, 2014.

\bibitem{dosovitskiy2015learning}
Alexey Dosovitskiy, Jost Tobias~Springenberg, and Thomas Brox.
\newblock Learning to generate chairs with convolutional neural networks.
\newblock In {\em CVPR}, 2015.

\bibitem{yang2015weakly}
Jimei Yang, Scott~E Reed, Ming-Hsuan Yang, and Honglak Lee.
\newblock Weakly-supervised disentangling with recurrent transformations for 3d
  view synthesis.
\newblock In {\em NIPS}, 2015.

\bibitem{kulkarni2015deep}
Tejas~D Kulkarni, William~F Whitney, Pushmeet Kohli, and Josh Tenenbaum.
\newblock Deep convolutional inverse graphics network.
\newblock In {\em NIPS}, 2015.

\bibitem{jimenez2016unsupervised}
Danilo Jimenez~Rezende, SM~Eslami, Shakir Mohamed, Peter Battaglia, Max
  Jaderberg, and Nicolas Heess.
\newblock Unsupervised learning of 3d structure from images.
\newblock {\em NIPS}, 2016.

\bibitem{yin2017towards}
Xi~Yin, Xiang Yu, Kihyuk Sohn, Xiaoming Liu, and Manmohan Chandraker.
\newblock Towards large-pose face frontalization in the wild.
\newblock In {\em ICCV}, 2017.

\bibitem{zhu2018visual}
Jun-Yan Zhu, Zhoutong Zhang, Chengkai Zhang, Jiajun Wu, Antonio Torralba, Josh
  Tenenbaum, and Bill Freeman.
\newblock Visual object networks: Image generation with disentangled 3d
  representations.
\newblock In {\em NeurIPS}, 2018.

\bibitem{nguyen2019hologan}
Thu Nguyen-Phuoc, Chuan Li, Lucas Theis, Christian Richardt, and Yong-Liang
  Yang.
\newblock Hologan: Unsupervised learning of 3d representations from natural
  images.
\newblock In {\em ICCV}, 2019.

\bibitem{nguyen2020blockgan}
Thu~H Nguyen-Phuoc, Christian Richardt, Long Mai, Yongliang Yang, and Niloy
  Mitra.
\newblock Blockgan: Learning 3d object-aware scene representations from
  unlabelled images.
\newblock 2020.

\bibitem{liao2020towards}
Yiyi Liao, Katja Schwarz, Lars Mescheder, and Andreas Geiger.
\newblock Towards unsupervised learning of generative models for 3d
  controllable image synthesis.
\newblock In {\em CVPR}, 2020.

\bibitem{schwarz2020graf}
Katja Schwarz, Yiyi Liao, Michael Niemeyer, and Andreas Geiger.
\newblock Graf: Generative radiance fields for 3d-aware image synthesis.
\newblock In {\em NeurIPS}, 2020.

\bibitem{chan2021pi}
Eric~R Chan, Marco Monteiro, Petr Kellnhofer, Jiajun Wu, and Gordon Wetzstein.
\newblock pi-gan: Periodic implicit generative adversarial networks for
  3d-aware image synthesis.
\newblock In {\em CVPR}, 2021.

\bibitem{hao2021gancraft}
Zekun Hao, Arun Mallya, Serge Belongie, and Ming-Yu Liu.
\newblock Gancraft: Unsupervised 3d neural rendering of minecraft worlds.
\newblock In {\em ICCV}, 2021.

\bibitem{zhou2021CIPS3D}
Peng Zhou, Lingxi Xie, Bingbing Ni, and Qi~Tian.
\newblock {{CIPS}}-{{3D}}: A {{3D}}-{{Aware Generator}} of {{GANs Based}} on
  {{Conditionally}}-{{Independent Pixel Synthesis}}.
\newblock 2021.

\bibitem{gu2022stylenerf}
Jiatao Gu, Lingjie Liu, Peng Wang, and Christian Theobalt.
\newblock Stylenerf: A style-based 3d aware generator for high-resolution image
  synthesis.
\newblock In {\em ICLR}, 2022.

\bibitem{or2022stylesdf}
Roy Or-El, Xuan Luo, Mengyi Shan, Eli Shechtman, Jeong~Joon Park, and Ira
  Kemelmacher-Shlizerman.
\newblock Stylesdf: High-resolution 3d-consistent image and geometry
  generation.
\newblock In {\em CVPR}, 2022.

\bibitem{poole2022dreamfusion}
Ben Poole, Ajay Jain, Jonathan~T Barron, and Ben Mildenhall.
\newblock Dreamfusion: Text-to-3d using 2d diffusion.
\newblock {\em arXiv preprint arXiv:2209.14988}, 2022.

\bibitem{skorokhodov2022epigraf}
Ivan Skorokhodov, Sergey Tulyakov, Yiqun Wang, and Peter Wonka.
\newblock Epigraf: Rethinking training of 3d gans.
\newblock 2022.

\bibitem{deng20233d}
Kangle Deng, Gengshan Yang, Deva Ramanan, and Jun-Yan Zhu.
\newblock 3d-aware conditional image synthesis.
\newblock In {\em CVPR}, 2023.

\bibitem{skorokhodov3d}
Ivan Skorokhodov, Aliaksandr Siarohin, Yinghao Xu, Jian Ren, Hsin-Ying Lee,
  Peter Wonka, and Sergey Tulyakov.
\newblock 3d generation on imagenet.
\newblock In {\em International Conference on Learning Representations}.

\bibitem{sargent2023vq3d}
Kyle Sargent, Jing~Yu Koh, Han Zhang, Huiwen Chang, Charles Herrmann, Pratul
  Srinivasan, Jiajun Wu, and Deqing Sun.
\newblock Vq3d: Learning a 3d-aware generative model on imagenet.
\newblock {\em arXiv preprint arXiv:2302.06833}, 2023.

\bibitem{wu20153d}
Zhirong Wu, Shuran Song, Aditya Khosla, Fisher Yu, Linguang Zhang, Xiaoou Tang,
  and Jianxiong Xiao.
\newblock 3d shapenets: A deep representation for volumetric shapes.
\newblock In {\em CVPR}, 2015.

\bibitem{wu2016learning}
Jiajun Wu, Chengkai Zhang, Tianfan Xue, Bill Freeman, and Josh Tenenbaum.
\newblock Learning a probabilistic latent space of object shapes via 3d
  generative-adversarial modeling.
\newblock 2016.

\bibitem{girdhar2016learning}
Rohit Girdhar, David~F Fouhey, Mikel Rodriguez, and Abhinav Gupta.
\newblock Learning a predictable and generative vector representation for
  objects.
\newblock In {\em ECCV}. Springer, 2016.

\bibitem{gadelha20173d}
Matheus Gadelha, Subhransu Maji, and Rui Wang.
\newblock 3d shape induction from 2d views of multiple objects.
\newblock In {\em 3DV}, 2017.

\bibitem{smith2017improved}
Edward~J Smith and David Meger.
\newblock Improved adversarial systems for 3d object generation and
  reconstruction.
\newblock In {\em CoRL}. PMLR, 2017.

\bibitem{henzler2019escaping}
Philipp Henzler, Niloy~J Mitra, and Tobias Ritschel.
\newblock Escaping plato's cave: 3d shape from adversarial rendering.
\newblock In {\em ICCV}, 2019.

\bibitem{lunz2020inverse}
Sebastian Lunz, Yingzhen Li, Andrew Fitzgibbon, and Nate Kushman.
\newblock Inverse graphics gan: Learning to generate 3d shapes from
  unstructured 2d data.
\newblock {\em arXiv preprint arXiv:2002.12674}, 2020.

\bibitem{zhou20213d}
Linqi Zhou, Yilun Du, and Jiajun Wu.
\newblock 3d shape generation and completion through point-voxel diffusion.
\newblock In {\em ICCV}, 2021.

\bibitem{ibing2021octree}
Moritz Ibing, Gregor Kobsik, and Leif Kobbelt.
\newblock Octree transformer: Autoregressive 3d shape generation on
  hierarchically structured sequences.
\newblock {\em arXiv preprint arXiv:2111.12480}, 2021.

\bibitem{achlioptas2018learning}
Panos Achlioptas, Olga Diamanti, Ioannis Mitliagkas, and Leonidas Guibas.
\newblock Learning representations and generative models for 3d point clouds.
\newblock In {\em ICML}. PMLR, 2018.

\bibitem{yang2019pointflow}
Guandao Yang, Xun Huang, Zekun Hao, Ming-Yu Liu, Serge Belongie, and Bharath
  Hariharan.
\newblock Pointflow: 3d point cloud generation with continuous normalizing
  flows.
\newblock In {\em ICCV}, 2019.

\bibitem{mo2020pt2pc}
Kaichun Mo, He~Wang, Xinchen Yan, and Leonidas Guibas.
\newblock {PT2PC}: Learning to generate 3d point cloud shapes from part tree
  conditions.
\newblock 2020.

\bibitem{sinha2017surfnet}
Ayan Sinha, Asim Unmesh, Qixing Huang, and Karthik Ramani.
\newblock Surfnet: Generating 3d shape surfaces using deep residual networks.
\newblock In {\em CVPR}, 2017.

\bibitem{groueix2018papier}
Thibault Groueix, Matthew Fisher, Vladimir~G Kim, Bryan~C Russell, and Mathieu
  Aubry.
\newblock A papier-m{\^a}ch{\'e} approach to learning 3d surface generation.
\newblock In {\em CVPR}, 2018.

\bibitem{pavllo2020convolutional}
Dario Pavllo, Graham Spinks, Thomas Hofmann, Marie-Francine Moens, and Aurelien
  Lucchi.
\newblock Convolutional generation of textured 3d meshes.
\newblock {\em NeurIPS}, 2020.

\bibitem{pavllo2021learning}
Dario Pavllo, Jonas Kohler, Thomas Hofmann, and Aurelien Lucchi.
\newblock Learning generative models of textured 3d meshes from real-world
  images.
\newblock In {\em ICCV}, 2021.

\bibitem{chen2019learningto}
Wenzheng Chen, Huan Ling, Jun Gao, Edward Smith, Jaakko Lehtinen, Alec
  Jacobson, and Sanja Fidler.
\newblock Learning to predict 3d objects with an interpolation-based
  differentiable renderer.
\newblock {\em NeurIPS}, 2019.

\bibitem{nash2020polygen}
Charlie Nash, Yaroslav Ganin, SM~Ali Eslami, and Peter Battaglia.
\newblock Polygen: An autoregressive generative model of 3d meshes.
\newblock In {\em ICML}. PMLR, 2020.

\bibitem{gao2022get3d}
Jun Gao, Tianchang Shen, Zian Wang, Wenzheng Chen, Kangxue Yin, Daiqing Li,
  Or~Litany, Zan Gojcic, and Sanja Fidler.
\newblock Get3d: A generative model of high quality 3d textured shapes learned
  from images.
\newblock In {\em NeurIPS}, 2022.

\bibitem{mo2019structurenet}
Kaichun Mo, Paul Guerrero, Li~Yi, Hao Su, Peter Wonka, Niloy Mitra, and
  Leonidas Guibas.
\newblock Structurenet: Hierarchical graph networks for 3d shape generation.
\newblock {\em ACM Transactions on Graphics (TOG), Siggraph Asia 2019},
  38(6):Article 242, 2019.

\bibitem{tulsiani2017learning}
Shubham Tulsiani, Hao Su, Leonidas~J Guibas, Alexei~A Efros, and Jitendra
  Malik.
\newblock Learning shape abstractions by assembling volumetric primitives.
\newblock In {\em CVPR}, 2017.

\bibitem{liu2019learning}
Shichen Liu, Shunsuke Saito, Weikai Chen, and Hao Li.
\newblock Learning to infer implicit surfaces without 3d supervision.
\newblock {\em NeurIPS}, 2019.

\bibitem{mescheder2019occupancy}
Lars Mescheder, Michael Oechsle, Michael Niemeyer, Sebastian Nowozin, and
  Andreas Geiger.
\newblock Occupancy networks: Learning 3d reconstruction in function space.
\newblock In {\em CVPR}, 2019.

\bibitem{luo2021surfgen}
Andrew Luo, Tianqin Li, Wen-Hao Zhang, and Tai~Sing Lee.
\newblock Surfgen: Adversarial 3d shape synthesis with explicit surface
  discriminators.
\newblock In {\em ICCV}, 2021.

\bibitem{chen2019learning}
Zhiqin Chen and Hao Zhang.
\newblock Learning implicit fields for generative shape modeling.
\newblock In {\em CVPR}, 2019.

\bibitem{yan2022shapeformer}
Xingguang Yan, Liqiang Lin, Niloy~J Mitra, Dani Lischinski, Daniel Cohen-Or,
  and Hui Huang.
\newblock Shapeformer: Transformer-based shape completion via sparse
  representation.
\newblock In {\em CVPR}, 2022.

\bibitem{shen2021deep}
Tianchang Shen, Jun Gao, Kangxue Yin, Ming-Yu Liu, and Sanja Fidler.
\newblock Deep marching tetrahedra: a hybrid representation for high-resolution
  3d shape synthesis.
\newblock {\em NeurIPS}, 2021.

\bibitem{mittal2022autosdf}
Paritosh Mittal, Yen-Chi Cheng, Maneesh Singh, and Shubham Tulsiani.
\newblock Autosdf: Shape priors for 3d completion, reconstruction and
  generation.
\newblock In {\em CVPR}, 2022.

\bibitem{richter2016playing}
Stephan~R Richter, Vibhav Vineet, Stefan Roth, and Vladlen Koltun.
\newblock Playing for data: Ground truth from computer games.
\newblock In {\em ECCV}. Springer, 2016.

\bibitem{hong2018learning}
Seunghoon Hong, Xinchen Yan, Thomas~S Huang, and Honglak Lee.
\newblock Learning hierarchical semantic image manipulation through structured
  representations.
\newblock {\em NeurIPS}, 2018.

\bibitem{kim2021drivegan}
Seung~Wook Kim, Jonah Philion, Antonio Torralba, and Sanja Fidler.
\newblock Drivegan: Towards a controllable high-quality neural simulation.
\newblock In {\em CVPR}, 2021.

\bibitem{li2019aads}
Wei Li, CW~Pan, Rong Zhang, JP~Ren, YX~Ma, Jin Fang, FL~Yan, QC~Geng, XY~Huang,
  HJ~Gong, et~al.
\newblock Aads: Augmented autonomous driving simulation using data-driven
  algorithms.
\newblock {\em Science robotics}, 4(28):eaaw0863, 2019.

\bibitem{ling2020variational}
Huan Ling, David Acuna, Karsten Kreis, Seung~Wook Kim, and Sanja Fidler.
\newblock Variational amodal object completion.
\newblock In {\em NeurIPS}, 2020.

\bibitem{zhang2021ners}
Jason Zhang, Gengshan Yang, Shubham Tulsiani, and Deva Ramanan.
\newblock Ners: Neural reflectance surfaces for sparse-view 3d reconstruction
  in the wild.
\newblock In {\em NeurIPS}, 2021.

\bibitem{zakharov2021single}
Sergey Zakharov, Rares~Andrei Ambrus, Vitor~Campagnolo Guizilini, Dennis Park,
  Wadim Kehl, Fredo Durand, Joshua~B Tenenbaum, Vincent Sitzmann, Jiajun Wu,
  and Adrien Gaidon.
\newblock Single-shot scene reconstruction.
\newblock In {\em CoRL}, 2021.

\bibitem{monnier2022share}
Tom Monnier, Matthew Fisher, Alexei~A Efros, and Mathieu Aubry.
\newblock Share with thy neighbors: Single-view reconstruction by
  cross-instance consistency.
\newblock In {\em ECCV}, 2022.

\bibitem{muller2022autorf}
Norman M{\"u}ller, Andrea Simonelli, Lorenzo Porzi, Samuel~Rota Bul{\`o},
  Matthias Nie{\ss}ner, and Peter Kontschieder.
\newblock Autorf: Learning 3d object radiance fields from single view
  observations.
\newblock In {\em CVPR}, 2022.

\bibitem{devaranjan2020meta}
Jeevan Devaranjan, Amlan Kar, and Sanja Fidler.
\newblock Meta-sim2: Unsupervised learning of scene structure for synthetic
  data generation.
\newblock In {\em ECCV}. Springer, 2020.

\bibitem{tan2021scenegen}
Shuhan Tan, Kelvin Wong, Shenlong Wang, Sivabalan Manivasagam, Mengye Ren, and
  Raquel Urtasun.
\newblock Scenegen: Learning to generate realistic traffic scenes.
\newblock In {\em CVPR}, 2021.

\bibitem{yang2020surfelgan}
Zhenpei Yang, Yuning Chai, Dragomir Anguelov, Yin Zhou, Pei Sun, Dumitru Erhan,
  Sean Rafferty, and Henrik Kretzschmar.
\newblock Surfelgan: Synthesizing realistic sensor data for autonomous driving.
\newblock In {\em CVPR}, 2020.

\bibitem{kundu2022panoptic}
Abhijit Kundu, Kyle Genova, Xiaoqi Yin, Alireza Fathi, Caroline Pantofaru,
  Leonidas~J Guibas, Andrea Tagliasacchi, Frank Dellaert, and Thomas
  Funkhouser.
\newblock Panoptic neural fields: A semantic object-aware neural scene
  representation.
\newblock In {\em CVPR}, 2022.

\bibitem{rematas2022urban}
Konstantinos Rematas, Andrew Liu, Pratul~P Srinivasan, Jonathan~T Barron,
  Andrea Tagliasacchi, Thomas Funkhouser, and Vittorio Ferrari.
\newblock Urban radiance fields.
\newblock In {\em CVPR}, 2022.

\bibitem{dosovitskiy2020image}
Alexey Dosovitskiy, Lucas Beyer, Alexander Kolesnikov, Dirk Weissenborn,
  Xiaohua Zhai, Thomas Unterthiner, Mostafa Dehghani, Matthias Minderer, Georg
  Heigold, Sylvain Gelly, et~al.
\newblock An image is worth 16x16 words: Transformers for image recognition at
  scale.
\newblock {\em arXiv preprint arXiv:2010.11929}, 2020.

\bibitem{hu2021unit}
Ronghang Hu and Amanpreet Singh.
\newblock Unit: Multimodal multitask learning with a unified transformer.
\newblock In {\em ICCV}, 2021.

\bibitem{jaegle2021perceiver}
Andrew Jaegle, Sebastian Borgeaud, Jean-Baptiste Alayrac, Carl Doersch, Catalin
  Ionescu, David Ding, Skanda Koppula, Daniel Zoran, Andrew Brock, Evan
  Shelhamer, et~al.
\newblock Perceiver io: A general architecture for structured inputs \&
  outputs.
\newblock {\em arXiv preprint arXiv:2107.14795}, 2021.

\bibitem{sajjadi2022scene}
Mehdi~SM Sajjadi, Henning Meyer, Etienne Pot, Urs Bergmann, Klaus Greff, Noha
  Radwan, Suhani Vora, Mario Lu{\v{c}}i{\'c}, Daniel Duckworth, Alexey
  Dosovitskiy, et~al.
\newblock Scene representation transformer: Geometry-free novel view synthesis
  through set-latent scene representations.
\newblock In {\em CVPR}, 2022.

\bibitem{rebain2022attention}
Daniel Rebain, Mark~J Matthews, Kwang~Moo Yi, Gopal Sharma, Dmitry Lagun, and
  Andrea Tagliasacchi.
\newblock Attention beats concatenation for conditioning neural fields.
\newblock {\em arXiv preprint arXiv:2209.10684}, 2022.

\bibitem{karras2020analyzing}
Tero Karras, Samuli Laine, Miika Aittala, Janne Hellsten, Jaakko Lehtinen, and
  Timo Aila.
\newblock Analyzing and improving the image quality of stylegan.
\newblock In {\em CVPR}, 2020.

\bibitem{peng2020convolutional}
Songyou Peng, Michael Niemeyer, Lars Mescheder, Marc Pollefeys, and Andreas
  Geiger.
\newblock Convolutional occupancy networks.
\newblock In {\em ECCV}. Springer, 2020.

\bibitem{shen2022acid}
Bokui Shen, Zhenyu Jiang, Christopher Choy, Leonidas~J Guibas, Silvio Savarese,
  Anima Anandkumar, and Yuke Zhu.
\newblock Acid: Action-conditional implicit visual dynamics for deformable
  object manipulation.
\newblock 2022.

\bibitem{mildenhall2021nerf}
Ben Mildenhall, Pratul~P Srinivasan, Matthew Tancik, Jonathan~T Barron, Ravi
  Ramamoorthi, and Ren Ng.
\newblock Nerf: Representing scenes as neural radiance fields for view
  synthesis.
\newblock In {\em ECCV}, 2020.

\bibitem{razavi2019generating}
Ali Razavi, Aaron Van~den Oord, and Oriol Vinyals.
\newblock Generating diverse high-fidelity images with vq-vae-2.
\newblock In {\em NeurIPS}, 2019.

\bibitem{chen2020generative}
Mark Chen, Alec Radford, Rewon Child, Jeffrey Wu, Heewoo Jun, David Luan, and
  Ilya Sutskever.
\newblock Generative pretraining from pixels.
\newblock In {\em ICML}. PMLR, 2020.

\bibitem{yu2021vector}
Jiahui Yu, Xin Li, Jing~Yu Koh, Han Zhang, Ruoming Pang, James Qin, Alexander
  Ku, Yuanzhong Xu, Jason Baldridge, and Yonghui Wu.
\newblock Vector-quantized image modeling with improved vqgan.
\newblock In {\em ICLR}, 2021.

\bibitem{yu2022scaling}
Jiahui Yu, Yuanzhong Xu, Jing~Yu Koh, Thang Luong, Gunjan Baid, Zirui Wang,
  Vijay Vasudevan, Alexander Ku, Yinfei Yang, Burcu~Karagol Ayan, et~al.
\newblock Scaling autoregressive models for content-rich text-to-image
  generation.
\newblock {\em arXiv preprint arXiv:2206.10789}, 2022.

\bibitem{villegas2022phenaki}
Ruben Villegas, Mohammad Babaeizadeh, Pieter-Jan Kindermans, Hernan Moraldo,
  Han Zhang, Mohammad~Taghi Saffar, Santiago Castro, Julius Kunze, and Dumitru
  Erhan.
\newblock Phenaki: Variable length video generation from open domain textual
  description.
\newblock {\em arXiv preprint arXiv:2210.02399}, 2022.

\bibitem{zhang2018unreasonable}
Richard Zhang, Phillip Isola, Alexei~A Efros, Eli Shechtman, and Oliver Wang.
\newblock The unreasonable effectiveness of deep features as a perceptual
  metric.
\newblock In {\em CVPR}, 2018.

\bibitem{kobayashi2022decomposing}
Sosuke Kobayashi, Eiichi Matsumoto, and Vincent Sitzmann.
\newblock Decomposing nerf for editing via feature field distillation.
\newblock {\em arXiv preprint arXiv:2205.15585}, 2022.

\bibitem{tschernezki2022neural}
Vadim Tschernezki, Iro Laina, Diane Larlus, and Andrea Vedaldi.
\newblock Neural feature fusion fields: 3d distillation of self-supervised 2d
  image representations.
\newblock {\em arXiv preprint arXiv:2209.03494}, 2022.

\bibitem{song2022nerfplayer}
Liangchen Song, Anpei Chen, Zhong Li, Zhang Chen, Lele Chen, Junsong Yuan,
  Yi~Xu, and Andreas Geiger.
\newblock Nerfplayer: A streamable dynamic scene representation with decomposed
  neural radiance fields.
\newblock {\em arXiv preprint arXiv:2210.15947}, 2022.

\bibitem{zhang2021datasetgan}
Yuxuan Zhang, Huan Ling, Jun Gao, Kangxue Yin, Jean-Francois Lafleche, Adela
  Barriuso, Antonio Torralba, and Sanja Fidler.
\newblock Datasetgan: Efficient labeled data factory with minimal human effort.
\newblock In {\em CVPR}, 2021.

\bibitem{caron2021emerging}
Mathilde Caron, Hugo Touvron, Ishan Misra, Herv{\'e} J{\'e}gou, Julien Mairal,
  Piotr Bojanowski, and Armand Joulin.
\newblock Emerging properties in self-supervised vision transformers.
\newblock In {\em ICCV}, 2021.

\bibitem{mei2022waymo}
Jieru Mei, Alex~Zihao Zhu, Xinchen Yan, Hang Yan, Siyuan Qiao, Yukun Zhu,
  Liang-Chieh Chen, Henrik Kretzschmar, and Dragomir Anguelov.
\newblock Waymo open dataset: Panoramic video panoptic segmentation.
\newblock In {\em ECCV}, 2022.

\bibitem{kingma2014adam}
Diederik~P Kingma and Jimmy Ba.
\newblock Adam: A method for stochastic optimization.
\newblock {\em arXiv preprint arXiv:1412.6980}, 2014.

\bibitem{heusel2017gans}
Martin Heusel, Hubert Ramsauer, Thomas Unterthiner, Bernhard Nessler, and Sepp
  Hochreiter.
\newblock Gans trained by a two time-scale update rule converge to a local nash
  equilibrium.
\newblock {\em NIPS}, 30, 2017.

\bibitem{radford2021learning}
Alec Radford, Jong~Wook Kim, Chris Hallacy, Aditya Ramesh, Gabriel Goh,
  Sandhini Agarwal, Girish Sastry, Amanda Askell, Pamela Mishkin, Jack Clark,
  et~al.
\newblock Learning transferable visual models from natural language
  supervision.
\newblock In {\em ICML}. PMLR, 2021.

\bibitem{lorensen1987marching}
William~E Lorensen and Harvey~E Cline.
\newblock Marching cubes: A high resolution 3d surface construction algorithm.
\newblock {\em ACM siggraph computer graphics}, 21(4):163--169, 1987.

\bibitem{verbin2022ref}
Dor Verbin, Peter Hedman, Ben Mildenhall, Todd Zickler, Jonathan~T Barron, and
  Pratul~P Srinivasan.
\newblock Ref-nerf: Structured view-dependent appearance for neural radiance
  fields.
\newblock In {\em CVPR}, 2022.

\bibitem{ba2016layer}
Jimmy~Lei Ba, Jamie~Ryan Kiros, and Geoffrey~E Hinton.
\newblock Layer normalization.
\newblock {\em arXiv preprint arXiv:1607.06450}, 2016.

\bibitem{barron2021mip}
Jonathan~T Barron, Ben Mildenhall, Matthew Tancik, Peter Hedman, Ricardo
  Martin-Brualla, and Pratul~P Srinivasan.
\newblock Mip-nerf: A multiscale representation for anti-aliasing neural
  radiance fields.
\newblock In {\em ICCV}, 2021.

\bibitem{szegedy2016rethinking}
Christian Szegedy, Vincent Vanhoucke, Sergey Ioffe, Jon Shlens, and Zbigniew
  Wojna.
\newblock Rethinking the inception architecture for computer vision.
\newblock In {\em Proceedings of the IEEE conference on computer vision and
  pattern recognition}, pages 2818--2826, 2016.

\bibitem{itseez2015opencv}
Itseez.
\newblock Open source computer vision library.
\newblock \url{https://github.com/itseez/opencv}, 2015.

\bibitem{trimesh}
{Dawson-Haggerty et al.}
\newblock trimesh.

\end{thebibliography}
}

\onecolumn
{\hspace{-5mm} \LARGE{\textbf{Appendix}\\
}}

\maketitle
\begin{appendices}
%\begin{abstract}
In this supplementary document, %we first describes additional video visualizations of datasets and models in Sec.~\ref{sec:vid}. 
we first describe in details our proposed dataset and the processing behind it in Sec.~\ref{sec:dataset}. Then, we discuss various implementation details including network architectures, evaluation metrics and conditional synthesis details in Sec.~\ref{sec:imp_details}. Next, we examine ablation of different loss terms, and evaluate stage 1 model's performance. Finally, we discuss baselines in more details in Sec.~\ref{sec:baseline} and showcase mesh extraction visualization results in Sec.~\ref{sec:mesh}.
%\end{abstract}

\begin{figure}[h]
    \centering
    \includegraphics[width=\columnwidth]{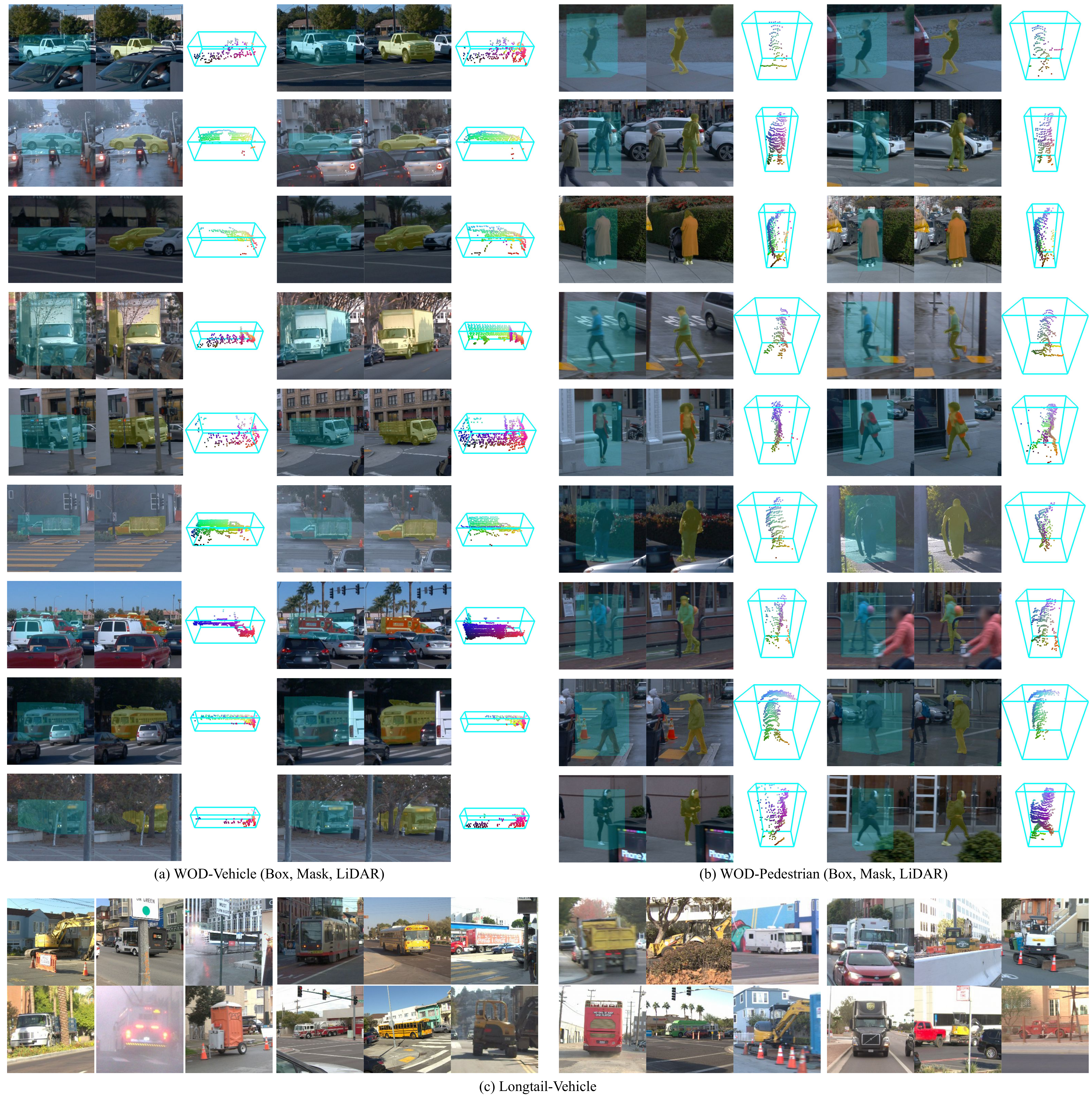}
    \cutcaptionup
    \cutcaptionup
    \caption{Our object-centric benchmark.}
    \cutcaptiondown
    \label{fig:dataset}
\end{figure}

\section{Dataset}
\label{sec:dataset}
We build the object-centric benchmark on top of the Waymo Open Dataset (WOD)~\cite{sun2020scalability} and our Longtail dataset.
The Waymo Open Dataset (WOD) is one of the largest and most diverse autonomous driving datasets among others~\cite{maddern20171,chang2019argoverse,caesar2020nuscenes}, containing rich geometric and semantic labels such as 3D bounding boxes and per-pixel instance masks.
Specifically, the dataset includes 1,150 driving scenes captured mostly in downtown San Francisco and Phoenix, each consisting of 200 frames of multi-sensor data.
Each data frame includes 3D point clouds from LiDAR sensors and high-resolution images from five cameras (positioned at Front, Front-Left, Front-Right, Side-Left, and Side-Right).
The objects were captured in the wild and their images exhibit large variations due to object interactions (e.g., heavy occlusion and distance to the robotic platform), sensor artifacts (e.g., motion blur and rolling shutter) and environmental factors (e.g., lighting and weather conditions).

To construct a benchmark for object-centric modeling, we propose a \textit{coarse-to-fine} procedure to extract collections of single-view 2D photographs by leveraging 3D object boxes, camera-LiDAR synchronization, and fine-grained 2D panoptic labels.
First, we leverage the 3D box annotations to exclude objects beyond certain distances to the surveying vehicle in each data frame (e.g., $40m$ for pedestrians and $60m$ for vehicles, respectively).
At a given frame, we project 3D point clouds within each 3D bounding box to the most visible camera and extract the centering patch to build our single-view 2D image collections.
Furthermore, we train a Panoptic-Deeplab model~\cite{cheng2020panoptic} using the 2D panoptic segmentations on the labeled subset~\cite{mei2022waymo} and create per-pixel \textit{pseudo-labels} for each camera image on the entire WOD.
This allows us to differentiate pixels belonging to the object of interest, background, and occluder (e.g., standing pole in front of a person).
We further exclude certain patches where objects are heavily occluded using the 2D panoptic predictions.
Even with the filtering criterion applied, we believe that the resulting benchmark is still very challenging due to occlusions, intra-class variations (e.g., truck and sedan), partial observations (e.g., we do not have full 360 degree observations of a single vehicle), and imperfect segmentation.
In particular, we provide accurate registration of camera rays and LiDAR point clouds to the object coordinate frame, taking into account the camera rolling shutter, object motion and ego motion.
Our \textit{WOD-ObjectAsset} can be accessed through \href{https://waymo.com/open/data/perception/#object-assets}{waymo.com/open}, organized in the Waymo Open Dataset modular format, enabling users to selectively download only the components they need.
Finally, we provide code examples to access and visualize data in the
\href{https://github.com/waymo-research/waymo-open-dataset/blob/master/tutorial/tutorial_object_asset.ipynb}{tutorial\_object\_asset}.

Our Longtail dataset contains LiDAR point clouds and camera images, along with 3D bounding box annotations.
We obtain the pseudo-labeled segmentations using the same 2D panoptic model pretrained on WOD.
We apply the same \textit{coarse-to-fine} procedure to obtain the Longtail-Vehicle benchmark.

\section{More Implementation Details}
\label{sec:imp_details}
\subsection{Network Architecture}
All models use  exponential moving average of weights.

\cutparagraphup
\paragraph{Encoder $E_\phi$.} 
Our encoder contains three vision transformer blocks and three cross-attention blocks. The vision transformer takes input images of resolution of $256^2$, and first map each patch into a $512$ dimensional token.
A \texttt{CLS} token is appended to the list of image patch tokens.
Then, the transformer blocks are used to process the image patch tokens.
Each transformer block has 8 heads, an embedding dimension of 512 and a hidden dimension of 2048. 
For cross-attention blocks, we first initialize tri-plane positional embedding of shape $16 \times 16 \times 3$, each embedding is of $512$ dimension.
The tri-plane positional embedding is passed through a fully-connected layer of $512$ dimension.
The processed tri-plane positional embedding is then used a query input to the cross-attention transformer blocks, while the image patch tokens serve as key and value.
Each cross-attention transformer block has 8 heads, an embedding dimension of 512 and a hidden dimension of 2048.
Finally, the output of the cross-attention transformer blocks are passed through a fully-connected layer with Layer Normalization~\cite{ba2016layer} and \texttt{tanh} activation into $16 \times 16  \times 3$ tokens of $32$ dimension, which is the dimension of each entry in the codebook $\mathbb{K}$.  

\cutparagraphup
\paragraph{Codebook $\mathbb{K}$.} Our discrete codebook contains 2048 entries with lookup dimension of 32, which means each entry is of 32-dimensional. Codebook are initialized using fan-in variance scaling, scale equals $1$ and uniform distribution. Similar to Yu \etal~\cite{yu2021vector}, we use $l_2$-normalized codes, which means applying $l_2$ normalization on the encoded tri-plane latents $\mathbf{e}^{\text{3D}}$ and codebook entries in $\mathbb{K}$.

\cutparagraphup
\paragraph{Decoder $G_\theta$ - Token Transformer.} The token transformer contains 3 self-attention transformers blocks. A \texttt{CLS} token is appended to the tri-plane latents. Positional encoding is used to represent 3D spatial locations. Each transformer block has 8 heads, an embedding dimension of 512 and a hidden dimension of 2048. Finally, the output of the transformer blocks are passed through a fully-connected layer with Layer Normalization~\cite{ba2016layer} and \texttt{tanh} activation into $16 \times 16  \times 3$ tokens of $256$ dimension (and an additional \texttt{CLS} token).

\cutparagraphup
\paragraph{Decoder $G_\theta$ - Style-based Generator.} We first use a mapping network~\cite{karras2020analyzing} to map the aforementioned \texttt{CLS} token into intermediate latent space $\mathbf{W}$. The mapping network contains 8 fully-connected layers of hidden dimension 512. The mapping network outputs a vector $w$ of $512$ dimensional. Following Karras \etal~\cite{karras2020analyzing}, we use $w$ for a style-based generator. For each plane in our tri-plane representation ($xy,xz,yz$ planes), we use a generator contains three up-sampling blocks with hidden dimensions of $512$, $256$ and $128$ respectively. Finally, the style-based generators output tri-plane feature maps with 32 feature channels.

\cutparagraphup
\paragraph{Decoder $G_\theta$ - Volume Rendering.} Our volume renderer is implemented as 2 fully-connected layers, similar to Chan~\etal\cite{chan2022efficient}. The decoder takes as input the 32-dimensional aggregated feature vector from the style-based generator. For each pixel, we query 40 points, with 24 uniformly sampled and 16 importance-sampled. We use MipNeRF~\cite{barron2021mip} as our volume rendering module. Volume rendering is performed at a resolution of $128\times 128$. 

\cutparagraphup
\paragraph{Discriminator.} We use a StyleGAN2~\cite{karras2020analyzing} discriminator with hidden dimensions $16,32,64,128,256$. We use R1 regularization with $\gamma=1$. 

\cutparagraphup
\paragraph{Stage-2 Modeling $M_\psi$.} We follow a shallower verions of the network architecture and training set up introduced in~\cite{chang2022maskgit}. We use $12$ layers, 8 attention heads, 768 embedding dimensions and 3072 hidden dimensions. The model uses learnable positional embedding, Layer Normalization, and truncated normal initialization (stddev$=0.02$). We use the following training hyperparameters: label smoothing=0.1, dropout rate=0.1, Adam optimizer~\cite{kingma2014adam} with $\beta_1=0.9$ and $\beta_2=0.96$. We use a cosine masking schedule. During inference, token synthesis are performed in $10$ steps. 

\subsection{Aligning Tri-plane to Object Scale}
\begin{figure*}[th]
\centering
\includegraphics[width=0.6\textwidth]{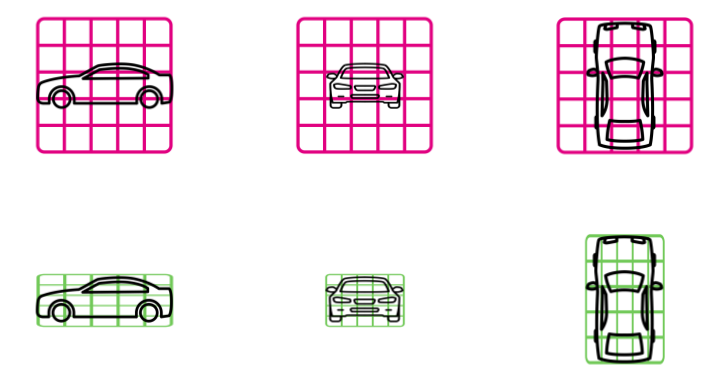}
\caption{Illustration of using uniform tri-plane versus using scale-aligned triplane. 
}
\cutcaptiondown
\label{fig:scaled_triplane}
\end{figure*}

Since vehicles can have drastically different scales in its $x,y,z$ directions, using a naive uniform scale tri-plane to cover the object leaves a lot of computation capacity under-utilized. As illustrated in the top row of Fig.~\ref{fig:scaled_triplane}, if we cover a normal sedan using uniform size tri-plane, most of the entries in the tri-plane features correspond to empty space. The problem becomes more severe for longer-tail instances of truck, bus \etc, where the scale ratio among $x,y,z$ become even more extreme. 

To encourage a more efficient tri-plane features usage, we make tri-plane latents aligned to object scales during the coordinate feature orthographic projection step. 
As illustrated in the bottom row of Fig.~\ref{fig:scaled_triplane}, when querying feature of coordinate $p\in[0,1]^3\subset\mathbb{R}^3$, if we have object scale ${s_x, s_y, s_z}$, we simply scale $p$ as $\hat{p}:=\frac{p}{[s_x,s_y,s_z]}\in[0,\frac{1}{s_x}]\times[0,\frac{1}{s_y}]\times[0,\frac{1}{s_z}]$, and query tri-plane features using $\hat{p}$.
The orthographic projection follows the same tri-plane grid-sampling and aggregation as in prior works~\cite{peng2020convolutional,shen2022acid,chan2022efficient}. 

In the basic GINA-3D pipeline without using scaled tri-plane features, the model learns to handle object scale implicitly. 
In our scaled box model variations, the model leverages the object scale only in tri-plane feature orthographic projection step.
The model implicitly learns to produce feature maps that align with object scale.
As illustrated in the main paper, such design greatly improve model performance. 
We leave feeding object scale information explicit to the model as a future direction to explore.

\subsection{Evaluation Metrics}
We discuss in details the metrics we have used for quantitative evaluations. 

\cutparagraphup
\paragraph{Image Quality.} To evaluate the image quality, we employ two metrics Fr\'echet Inception Distance (FID)~\cite{heusel2017gans} and Mask Floater-Over Union (Mask FOU) over 50K generated images.
Fr\'echet Inception Distance (FID)~\cite{heusel2017gans} is commonly used to evaluate the quality of 2D images. 
The generated images are encoded using a pretrained Inception v3~\cite{szegedy2016rethinking} model, and the last pooling layer's output was stored as the final encoding. The FID metric is computed as:
\begin{align}
\text{FID}(I_g, I_v) = ||\mu_g - \mu_v||_2^2 + \text{Tr}[\Sigma
_g + \Sigma_v - 2\sqrt{\Sigma
_g \cdot \Sigma_v }]
\end{align}
where Tr denotes the trace operation, $\mu_g,\Sigma_g$ are the mean and covariance matrix of the generated images encodings, and $\mu_v,\Sigma_v$ are the mean and covariance matrix of the validation images encodings. 

We additionally measure if the generated texture forms a single full object, which is implemented by checking if the generated pixels span a connected region.
We measure this by calculating percentage of pixels that are not connected.
Since all images from baselines and GINA-3D are generated using a white background, we measure pixels connected components using the \texttt{findContours} function from OpenCV~\cite{itseez2015opencv} to find connected components, and use \texttt{contourArea} to find the largest connected component, which we denote $C_l$. We then use the aggregated density (alpha) value to find the entire shape's projection on the image, which we denote $S$. 
Mask FOU is simply calculated mean over entire generated image set (as percentage):
\begin{align}
\text{Mask FOU}(I_g) = \frac{1}{|I_g|} \sum_{i\in I_g}(1-\frac{\text{Area}(C_{l,i})}{\text{Area}(S_{i})})
\end{align}

\cutparagraphup
\paragraph{Image Diversity.} We want to evaluate the semantic diversity of the generated image, which we measure with Coverage (COV) score and Minimum Matching Distance (MMD)~\cite{achlioptas2018learning} using pretrained CLIP~\cite{radford2021learning} embeddings. Specifically, Coverage (COV) score measures the fraction of images in the validation set that are matched to at least one of the images in the generated set. Formally, it's defined as:
\begin{align}
\text{COV}(I_g, I_v) = \frac{|\{\argmin_{i\in I_v} ||\text{CLIP}(i)-\text{CLIP}(j)||_2^2|j \in I_g\}|}{|I_v|}
\end{align}
Intuitively, COV uses CLIP embedding distance to perform nearest-neighbor matching for each generate image towards validation set. It measures diversity by checking what percentage of validation set is being matched as a nearest neighbor. However, COV is only one side of the story. A set of generated image can have a high COV score by having purely random generated images that are randomly matched to validation set. This issue is alleviated by the incorporation of Minimum Matching Distance (MMD), which measures if the nearest-neighbor matching yields high-quality matching pairs:
\begin{align}
\text{MMD}(I_g, I_v) = \frac{1}{|I_v|}\sum_{i\in I_v} \min_{j\in I_g} ||\text{CLIP}(i)-\text{CLIP}(j)||_2^2 
\end{align}
Intuitively, MMD measures the average closest distance between images in the validation set and their corresponding nearest neighbor in the training set. MMD correlates well with how faithful (with respect to the validation set) elements of generated set are~\cite{achlioptas2018learning}.

\cutparagraphup
\paragraph{Geometry Quality.} Due to a lack of 3D geometry ground-truth for in-the-wild data, we measure geometry quality using an existing metric Consistency score from Or-El \etal~\cite{or2022stylesdf}, and a Mesh Floater-Over Union (Mesh FOU) which measures if the geometry forms a single connected object.
Consistency score measures if the implicit fields are evaluated at consistent 3D locations, which is an important characteristic for view-consistent renderings~\cite{or2022stylesdf}.
In practice, it measures depth map consistency across viewpoints by back-projecting depth map to the 3D space.
For each model, we normalize the object longest edge to length of $10$ for numeric clarity, and compare two depth maps at an angle difference of $45$ degrees along the $z$-axis (yaw).
We calculate consistency across depth maps for all images in the generated set, denote as $D_g$:
\begin{align}
\text{Consistency}(D_g) = \frac{1}{|D_g|}\sum_{i\in D_g}\text{CD}(i, i_{rot})
\end{align}
where $i_{rot}$ represents the depth map after rotating the view point by 45 degree along $z$-axis.

We additionally measure if each generated shape forms a single full object, which is measured by checking if the generated mesh forms a single mesh.
We measure this by calculating percentage of mesh surface area that is not connected. 
We use surface area over volume because we observe that volume calculation is unstable with non-watertight meshes.
For each generated mesh $S$, we use \texttt{split} function from Trimesh~\cite{trimesh} to find the largest connected component, which we denote $C_l$. Mesh FOU is simply calculated mean over entire generated mesh set $M_g$ (as percentages):
\begin{align}
\text{Mesh FOU}(M_g) = \frac{1}{|M_g|} \sum_{i\in M_g}(1-\frac{\text{Area}(C_{l,i})}{\text{Area}(S_{i})})
\end{align}

\begin{figure*}[th]
\centering
\includegraphics[width=0.6\textwidth]{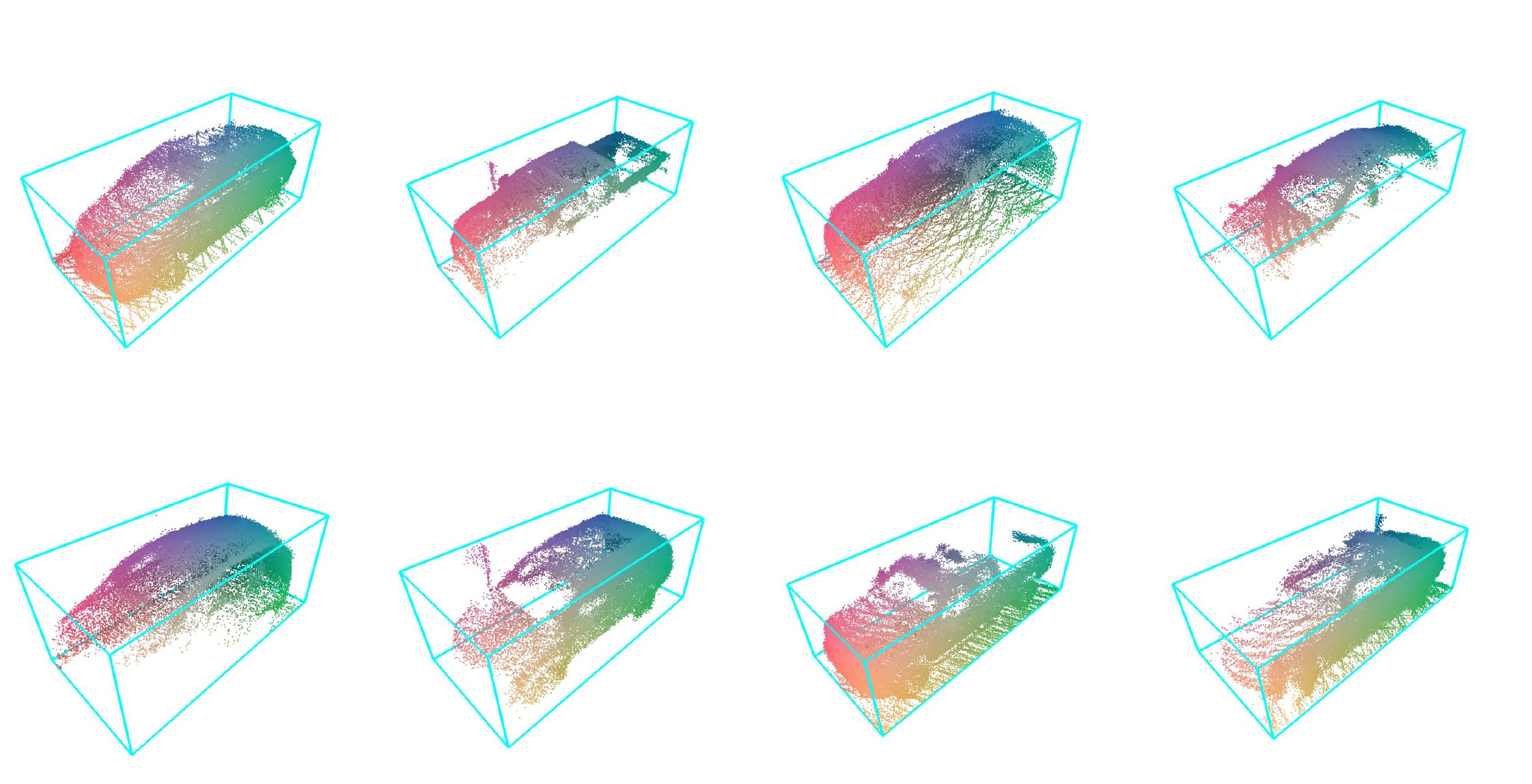}
\caption{Illustrations of aggregated point clouds.}
\cutcaptiondown
\label{fig:val_pcloud}
\end{figure*}
\cutparagraphup
\paragraph{Geometry Diversity.} We use Coverage (COV) and Minimum Matching Distance (MMD) again for measuring diversity. 
However, due to the lack of ground truth full 3D shape from in-the-wild data, our metric needs to be more carefully designed. 
A source for accurate but partial geometry that we can obtain is by aggregating LiDAR point-cloud scans for a given instance from different observations.
We then uniformly subsample 2048 points from the aggregated point cloud.
We show examples of aggregated point clouds in Fig.~\ref{fig:val_pcloud}.
As shown in the figure, the aggregated point clouds are indicative of the underlying shapes, but are incomplete. Chamfer distance, a common metric for shape similarity, calculates bi-directional nearest neighbors. However, due to incompleteness, finding the nearest neighbors of the generated points in the partial point will only result in noisy matches. 
Therefore, we do not measure the two-sided Chamfer distance, but measure only the distance of nearest neighbors of validation point clouds in the generated mesh. Formally, we have:
\begin{align}
\text{COV}(M_g, P_v) &= \frac{|\{\argmin_{i\in P_v} D(i,j) |j \in M_g\}|}{|P_v|} \\
\text{MMD}(M_g, P_v) &= \frac{1}{|P_v|}\sum_{i\in P_v} \min_{j\in M_g} D(i,j) \\
D(i,j | i\in P_v, j \in M_g) &= \frac{1}{|i|}\sum_{x\in i} \min_{y \in j} ||x-y||_2^2
\end{align}

\subsection{Conditional Synthesis}
We showcased in the main paper various conditional synthesis tasks, for which we provide more details here. 

\cutparagraphup
\paragraph{Discrete Conditions.} 
We feed discrete conditions (object class, time-of-day) as additional tokens to MaskGIT. Specifically, we increase the vocabulary size by the number of classes in the discrete conditions. Object class contains 4 options: cars, truck, bus and others. Time-of-day is a binary variable of day versus night. The vocabulary thus becomes $2048+4$ for object class, and $2048+2$ for time-of-day. We feed the conditional input as an additional token to the $768$ tri-plane latents by concatenating the two, resulting in an input of sequence length $769$. The sequence is then fed into MaskGIT for masked token prediction as in unconditional case.

\cutparagraphup
\paragraph{Continuous Conditions.} 
Alternatively, we feed continuous conditions to MaskGIT by concatenating conditional input with MaskGIT intermediate layer's output. Specifically, MaskGIT first generates word embedding for each token in the sequence. We pass the continuous condition through a fully-connected layer and concatenate the output with each token's word embedding. The concatenated embedding is then passed through the rest of the network. 
To synthesize samples conditioned on object semantics, we feed semantic embedding from a pre-trained DINO model~\cite{caron2021emerging}.
%pass in pre-trained object embedding like the \texttt{CLS} token from DINO~\cite{caron2021emerging}, which is of $768$ dimensional. 
%
To condition on object scale, we pass in positional embedding of object scale. We use standard cosine and sine positional embedding of degree 6. 

\cutparagraphup
\paragraph{Image-conditioned Assets Variations.} 
Given our mask-based iterative sampling stage, we can generate image-conditioned asset with variations.
We first use stage-1 model to perform reconstruction, retrieving a full-set of predicted tri-plane latents.
We then generate variations of the reconstructed instance by randomly masking out tri-plane latents.
The degree of variations can be controlled by masking out different number of  tokens.
By masking $90\%$ of tokens, we observe the variations are mostly reflected in generated assets under different textures.
By masking out $99\%$ of tokens, we see changes in object shapes more significantly, while the general object class remain the same.
We believe how to better control the variation process is an interesting direction to explore in the future.
\section{Additional GINA Visualizations}
\begin{figure}[t]
    \centering
    \includegraphics[width=0.75\columnwidth]{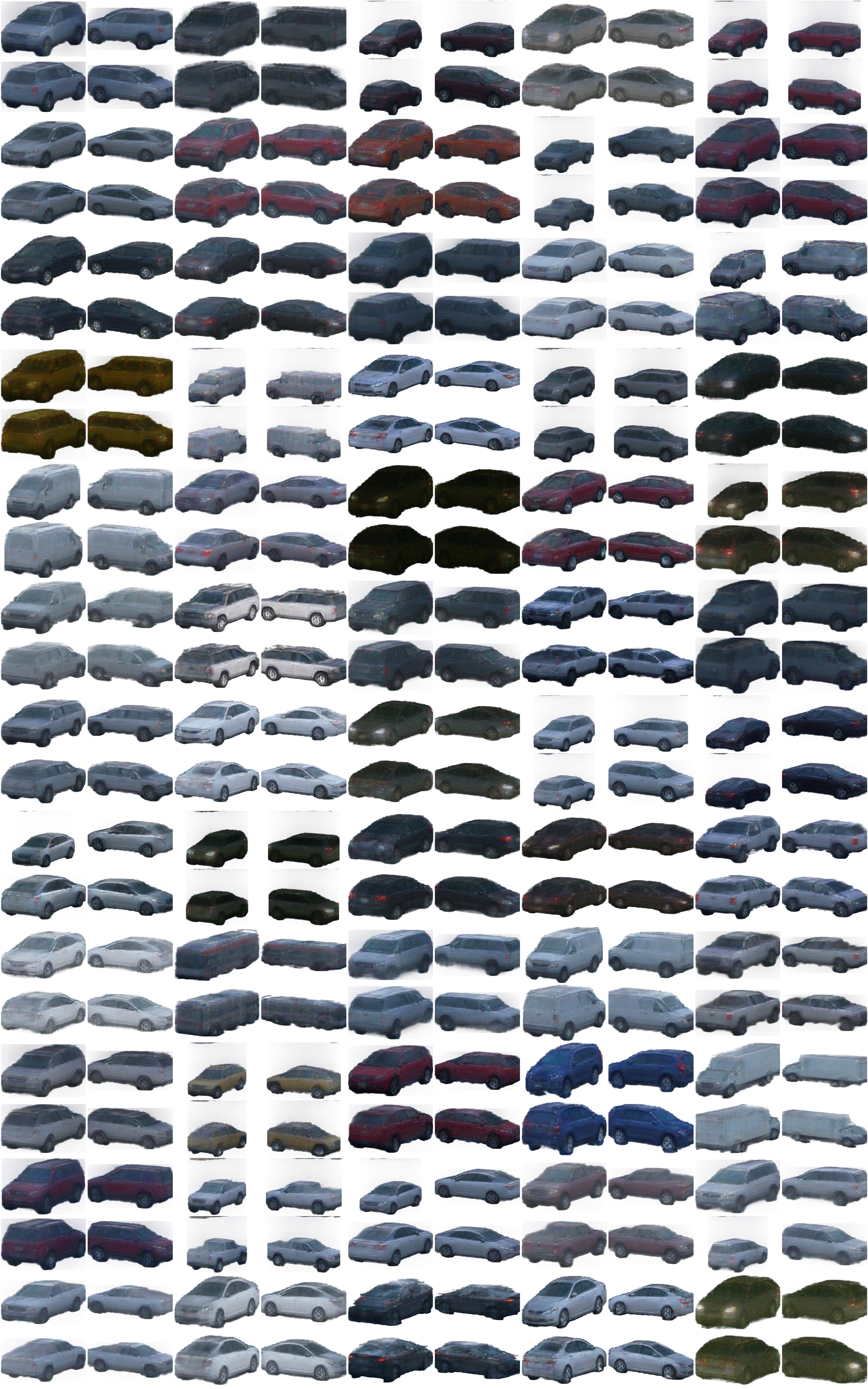}
    \caption{Additional qualitative results of GINA-3D.}
    \cutcaptiondown
    \label{fig:more_gina}
\end{figure}

We present additional visualizations of GINA-3D model in Fig.~\ref{fig:more_gina}.

\section{Ablation on Loss Terms and Stage 1 Evaluation}
\paragraph{Ablation study.} 
In this experiment, we use our scaled box model, trained with LiDAR supervision as our base model.
We conduct ablation studies by removing each loss, removing quantization entirely and training with different codebook sizes.
As shown in Table~\ref{tab:ablations}, the ablaion results justify each loss term we introduced in the paper, as removing each one of them leads to higher FID compared to the full model.
This finding is consistent with Esser \etal~\cite{esser2021taming}, which suggests LPIPS is important for visual fidelity. 
In addition, larger codebook $\mathbb{K}$ ($2^{12}$) has marginal impact in our setting. 

\begin{table}[t]
\centering
\begin{tabular}{l|cccc|c|cc|c}
\hline
{{\color{orange}1st}, {\color{teal}2nd}, {\color{cyan}3rd}}
                     & \sout{$\mathcal{L}_\text{GAN}$} 
                     & \sout{LPIPS} 
                     & \sout{$\mathcal{L}_\alpha$} 
                     & \sout{$\mathcal{L}_\text{VQ}$} 
                     & {No VQ} 
                     & {$|\mathbb{K}| = 2^{10}$} 
                     & {$|\mathbb{K}| = 2^{12}$} 
                     & {Full} \\ \hline
{Generative Metric (FID)}
& {65.1}   
& {83.0} 
& {80.3}  
& {\color{cyan}64.7}  
& -  
& {66.2}   
& {\color{orange}58.9}   
& {\color{teal}59.5}   
\\ \hline
Recon. Metric ($\ell_2$ {input view})
& {1.78}   
& {1.92} 
& {\color{teal}1.44}  
& {1.62}  
& {\color{orange}1.01}  
& {2.21}   
& {1.81}
& {\color{cyan}1.55}   
\\
Recon. Metric  ($\ell_2$ {cross view})
& {2.42}   
& {2.28} 
& {\color{orange}1.55}  
& {2.14}  
& {\color{teal}1.71}  
& {2.28}   
& {2.30}
& {\color{cyan}1.83}
\\
\hline
\end{tabular}
\caption{We perform various ablation studies on 1) removing each term in out overall loss function; 2) Removing vector quantization entirely; 3) Different codebook sizes $\mathbb{K}$. We further report stage 1 model's reconstruction quality using $\ell_2$ losses for input views as well as novel views.}
\label{tab:ablations}
\end{table}

\paragraph{Evaluating stage 1 model.} We report $\ell_2$ reconstruction loss (in $10^{-2}$) on the input and novel views of unseen instances. The model is able to obtain better reconstruction performance by removing quantization entirely (\textit{No VQ}), but it deprives the discrete codebook for stage 2 generative training. 
While generation and reconstruction correlates to some extent, performance rankings (color-coded) differ between them.

\begin{figure*}[t]
\centering
\includegraphics[width=0.95\linewidth]{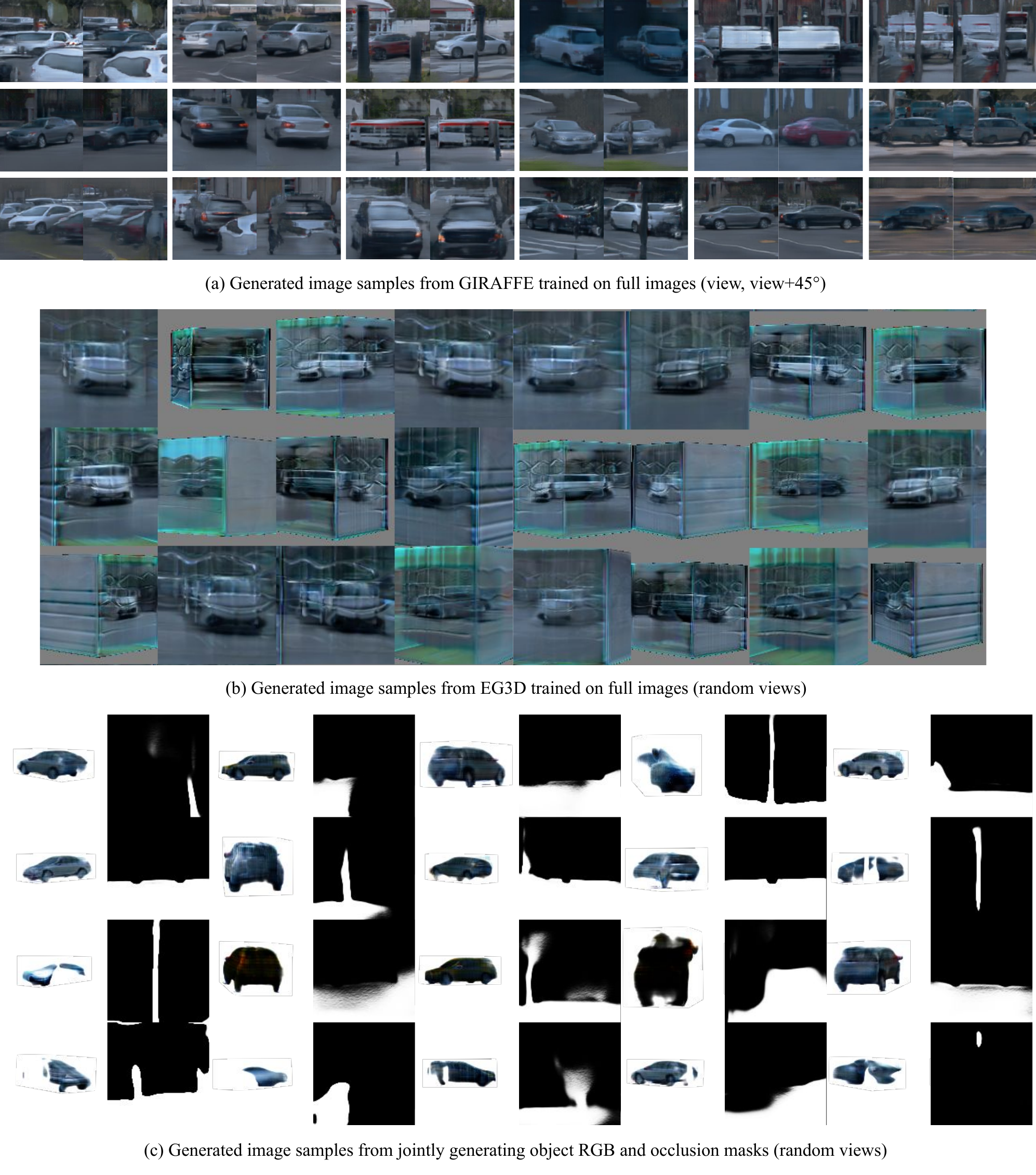}
\caption{Additional results on generation results trained on full images. a) We trained GIRAFFE on full images; b) We trained EG3D models on full images; c) We augmented the EG3D model by jointly generating object RGB, background RGB and occlusion masks. We visualizes object RGB and its corresponding occlusion mask in alternate columns. Results suggest that it's difficult for the model to disentangle object shape and occlusion.  
}
\label{fig:supp_gen_with_full_image}
\end{figure*}

\begin{figure}[t]
    \centering
    \includegraphics[width=\columnwidth]{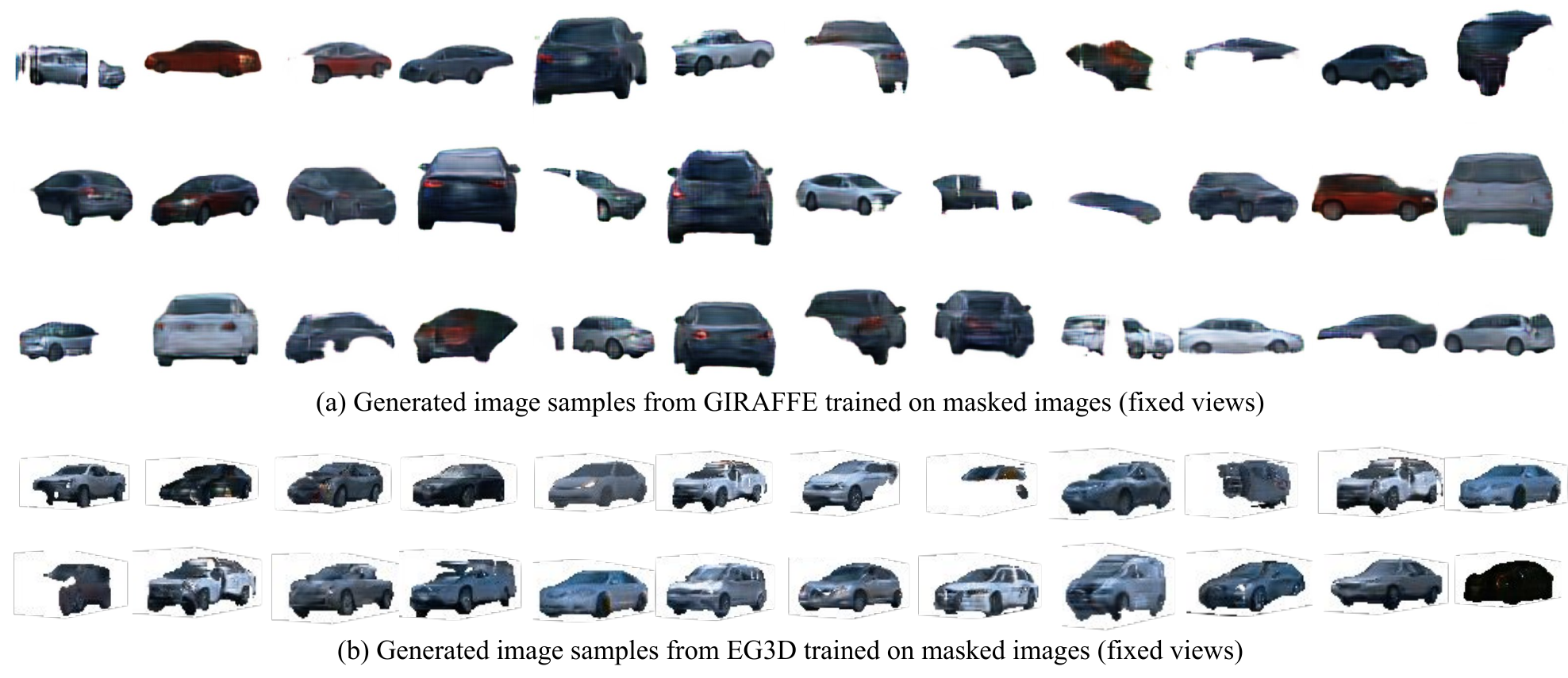}
    \cutcaptionup
    \cutcaptionup
    \caption{Additional results on generation results trained on masked images. a) Additional visualizations of GIRAFFE baseline reported in the main paper; b) Additional visualizations of EG3D baseline reported in the main paper. Results suggest that it's difficult for GIRAFFE to disentangle rotation. Both baselines show significant occlusion artifacts.}
    \cutcaptiondown
    \label{fig:more_baseline_results}
\end{figure}

\begin{figure*}[th]
\centering
\begin{subfigure}{.5\textwidth}
  \centering
  \includegraphics[width=0.95\linewidth]{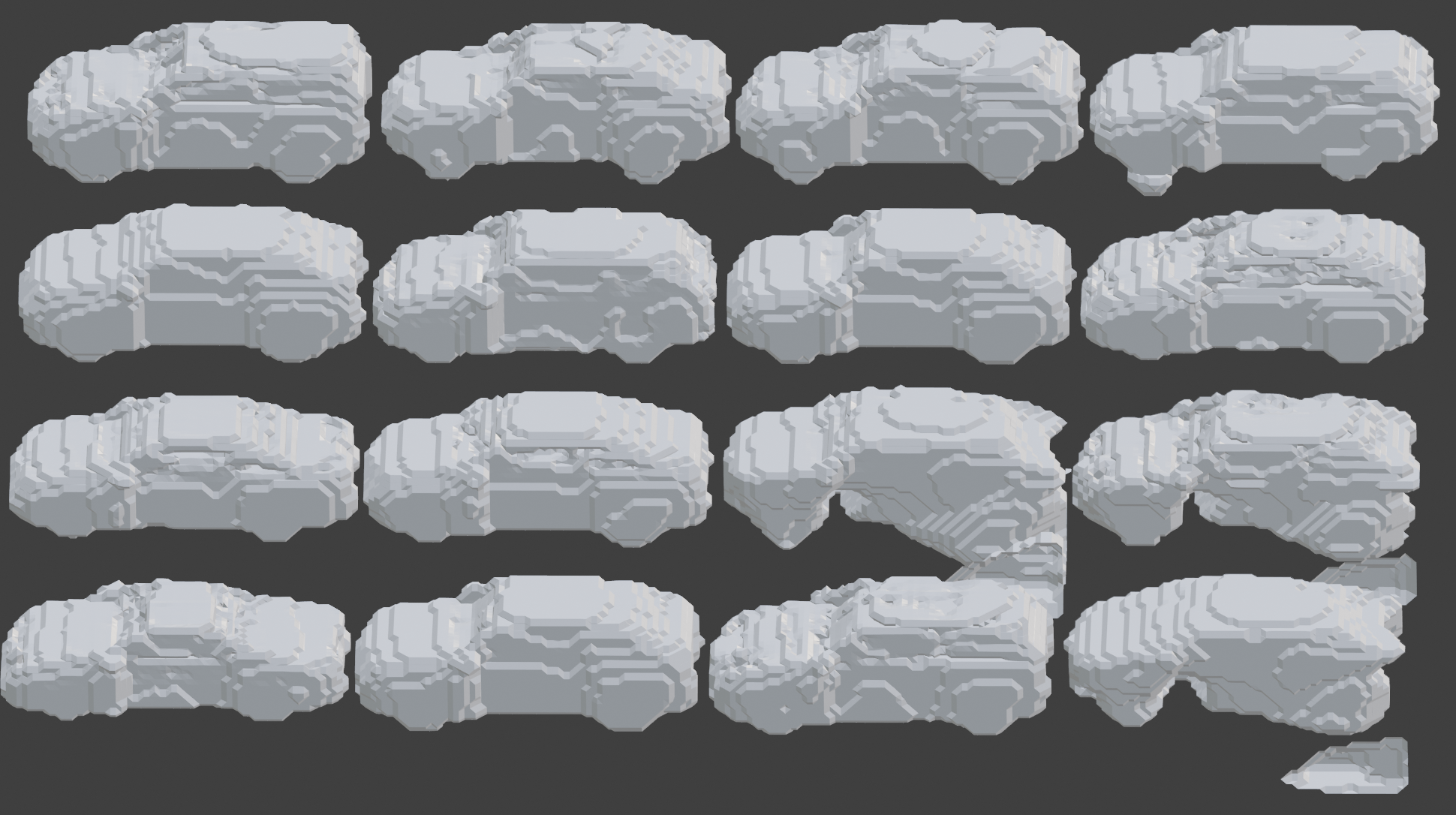}
  \caption{A random batch of 16 EG3D extracted meshes.}
  \label{fig:mesh_eg3d}
\end{subfigure}%
\begin{subfigure}{.5\textwidth}
  \centering
  \includegraphics[width=\linewidth]{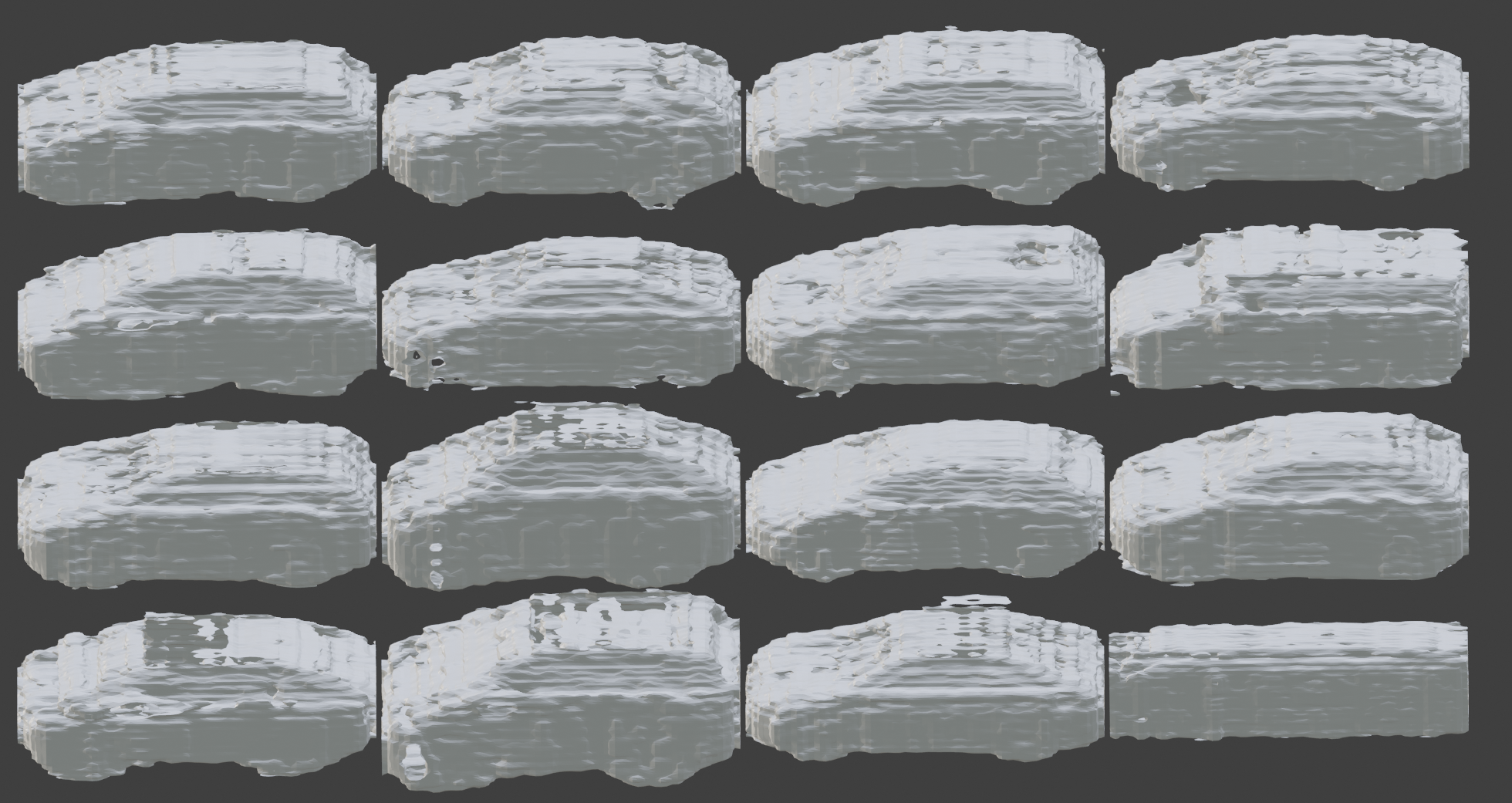}
  \caption{A random batch of 16 GINA-3D extracted meshes.}
  \label{fig:mesh_gina3d}
\end{subfigure}
\caption{Example mesh extractions from EG3D and GINA-3D.}
\label{fig:mesh_comp}
\end{figure*}

\section{Discussions on Baselines}
\label{sec:baseline}

%\cutparagraphup
\paragraph{Generating the full images.}
Directly modeling full images of data-in-the-wild yields significant challenges.
In the early stage of the project, we experimented with directly using GAN-based approaches on full images.
As illustrated in Fig.~\ref{fig:supp_gen_with_full_image}-a,b, feeding full images without explicit modeling of occlusion makes learning challenging on our benchmark.
For EG3D, we observed that training EG3D with unmasked image leads to training collapse, due to the absence of foreground and background modeling. 
For example, the generated image samples in Fig.~\ref{fig:more_baseline_results}-b lack diversity in shape and appearance (\eg color).

We clarify a key difference to pure GAN-based approaches (GIRAFFE and EG3D) is that our approach has two training stages and the masked loss is only applied in the first stage to \textit{reconstruct} the input.
In other words, masked loss cannot be directly applied to existing GAN-based approaches as the corresponding object mask for each generated RGB is not observed in the adversarial (encoder-free) training.
Alternatively, one can still apply the masked loss by factorizing RGB, object silhouette and occlusion. We have tried many variations of this idea in the early stage without avail, as learning disentangled factors was challenging for adversarial training. 
We provide such examples in Fig.~\ref{fig:supp_gen_with_full_image}-c. In this experiment, we tried to extend EG3D by generating occlusion masks with a separate branch.
However, the training became very unstable and we were not able to produce improved results beyond the original EG3D on our benchmark.
As we can see, the model fails to disentangle object silhouette and occlusion. It still generates partial shape, while generating some plausible foreground occlusion.
In fact, occlusion is even more challenging to generate explicitly on our data where object silhouettes and occlusion masks are entangled, as the outcome depends on the view and layout.

\cutparagraphup
\paragraph{Generating the object images.}

Whitening out non-object regions has been used by EG3D (see ShapeNet-Cars in its supp.) and GET3D. It combines white color to pixels with $\alpha < 1$ during neural rendering, which implicitly supervise $\alpha$. Such set up separates object pixels from the surroundings, and makes generation focused on object modeling.
We follow this design and have found in our experiments that baselines fail to generate separated target object without whitening-out.

We provide additional details about the baseline methods GIRAFFE and EG3D in Fig.~\ref{fig:more_baseline_results}.
We noticed that the learned GIRAFFE models are capable of generating vehicle-like patches but with viewpoints, occlusions and identities entangled in the latent space.
For example, we generate a pair of images (in Fig.~\ref{fig:more_baseline_results}(a)) by varying the viewpoint variable while keeping the identity latent variable fixed.
It turns out that the generations are not easily controllable by the viewpoint variables, while the vehicle identities often change across views.
The entangled representation makes the extracted meshes not very meaningful for the GIRAFFE baseline on our benchmark.
Additionally, the geometry extraction becomes even harder as the rendering mask is defined at a low dimensional resolution $16^2$.

\section{Extracted Meshes}
\label{sec:mesh}
As mentioned in the main text, we use marching cubes~\cite{lorensen1987marching} with density threshold of 10 to extract meshes for geometry evaluation. We showcase here random samples of extracted meshes from EG3D and GINA-3D. We show 16 examples each in Fig.~\ref{fig:mesh_eg3d}-\ref{fig:mesh_gina3d}. As we see, EG3D meshes can contain artifacts like missing parts of shape (row 3 right two). Furthermore, it shows relatively little diversity. GINA-3D not only preserves complete shapes, but also demonstrate a greater diversity, including more shape variation and semantic variation (mini-van row 2 column 4; bus row 4 column 4). Such observation is consistent with our quantitative evaluations. 

However, we do observe that GINA-3D meshes can be non-watertight and contain holes. We hope to address such problems in future works. We believe that by incorporating other representations like Signed Distance Fields (SDF), the mesh quality can be further improved. 

\end{appendices}

\end{document}